\documentclass[journal]{IEEEtran}
\usepackage{multirow}
\usepackage{color}
\usepackage{amsmath}
\usepackage{array}
\usepackage{stfloats}
\usepackage{url}
\usepackage{ragged2e}
\usepackage{cite}
\usepackage{amssymb}
\usepackage{xcolor}
\usepackage{bm}
\usepackage{algorithmicx,algorithm}
\usepackage[noend]{algpseudocode}
\usepackage{threeparttable}
\usepackage{threeparttable}
\usepackage{multirow}
\usepackage{booktabs}
\usepackage{pifont}
\usepackage{comment}
\usepackage{hyperref}
\newcommand{\tabincell}[2]{\begin{tabular}{@{}#1@{}}#2\end{tabular}}  
\newcommand{\cmark}{\ding{51}}%
\newcommand{\xmark}{\ding{55}}%

\ifCLASSINFOpdf
\usepackage[pdftex]{graphicx}
\graphicspath{{./figurepdf/}}
\DeclareGraphicsExtensions{.pdf}
\else
\usepackage[dvips]{graphicx}
\graphicspath{{./figureeps/}}
\DeclareGraphicsExtensions{.eps}
\fi
\begin{document}

\title{Adaptive Linear Span Network for Object Skeleton Detection}

\author{Chang Liu,
        Yunjie Tian, 
        Jianbin Jiao,~\IEEEmembership{Member,~IEEE,}
        Qixiang Ye,~\IEEEmembership{Senior Member,~IEEE}
\thanks{C. Liu, Y. Tian, J. Jiao, and Q. Ye are with the School of Electronic, Electrical, and Communication Engineering, University of Chinese Academy of Sciences, Beijing, 100049, China. E-mail: \{liuchang615@mails.ucas.ac.cn, tianyunjie19@mails.ucas.ac.cn, jiaojb@ucas.ac.cn, qxye@ucas.ac.cn\}.}
\thanks{C. Liu and Y. Tian contribute equally to this work.}
}

% The paper headers
\markboth{IEEE Transactions on Image Processing, Manuscript for Review}%
{Shell \MakeLowercase{\textit{et al.}}: Bare Demo of IEEEtran.cls for IEEE Journals}

\maketitle

\begin{abstract}
%指出传统方法的问题
Conventional networks for object skeleton detection are usually hand-crafted. 
%说明仍然没有自动的配置多尺度特征的方法
Although effective, they require intensive priori knowledge to configure representative features for objects in different scale granularity. 
%本文提出了AdaLSN，自动的配置多尺度的特征用于物体骨架检测
In this paper, we propose adaptive linear span network (AdaLSN), driven by neural architecture search (NAS), to automatically configure and integrate scale-aware features for object skeleton detection. 
%理论基础是Linear Span，这是最早的对特征融合的理论解释之一
AdaLSN is formulated with the theory of linear span, which provides one of the earliest explanations for multi-scale deep feature fusion. 
%通过定义unit与pyrmaid混合的空间实现，超出了很多现有的空间定义
AdaLSN is materialized by defining a mixed unit-pyramid search space, which goes beyond many existing search spaces using unit-level or pyramid-level features.
%在定义的空间内，遗传结构搜所自适应的优化配置两个层次的网络操作
Within the mixed space, we apply genetic architecture search to jointly optimize unit-level operations and pyramid-level connections for adaptive feature space expansion. 
AdaLSN substantiates its versatility by achieving significantly higher accuracy and latency trade-off compared with state-of-the-arts. 
It also demonstrates general applicability to image-to-mask tasks such as edge detection and road extraction. 
Code is available at \href{https://github.com/sunsmarterjie/SDL-Skeleton}{\color{magenta}github.com/sunsmarterjie/SDL-Skeleton}.

\end{abstract}


\IEEEpeerreviewmaketitle

\section{Introduction}
\IEEEPARstart{S}{keleton} is a kind of representative visual descriptor, which contains rich information about object topology, constituting an explicated abstraction of object shape. Object skeletons can be converted to descriptive features and/or spatial constraints, promoting computer vision tasks such as human pose estimation~\cite{pose2016}, hand gesture recognition~\cite{2DCurvedReflection2015}, text detection~\cite{SymmetryText2015}, and object localization~\cite{Proposal2015}, in an explainable fashion.

In the deep learning era, object skeleton detection using convolutional neural networks (CNNs) has made unprecedented progress. State-of-the-art approaches commonly utilize side-output architectures to integrate feature pyramid as a countermeasure for variations from object appearances, poses, and scales. This is intrinsically based on the observation that low-level features focus on detailed structures while high-level features are rich in semantics~\cite{RSRN2017}. 

As a pioneer work, the holistically-nested edge detection (HED)~\cite{HED2015} used a deeply supervised strategy to take full use of the hierarchical multi-scale features in a parallel manner. Fusing scale-associated deep side-outputs (FSDS)~\cite{FSDS2016} adopted a divided-and-conquer approach to supervise network side-outputs given scale-associated ground-truth. SRN~\cite{SRN2017} and RSRN~\cite{RSRN2017} investigated the multi-layer association problem by utilizing side-output residual units to pursue the complementarity among multi-scale features in a deep-to-shallow fasion. HiFi~\cite{HiFi2018} introduced a bilateral feature integration mechanism to incorporate the low-level details and high-level semantics.
 
Despite the encouraging progress, one limitation lies in that most of the existing network architectures for skeleton detection are hand-crafted and lack the theoretical backup. Although such networks incorporating rich human knowledge are somewhat effective, they experience difficulty in maximizing representation complementarity for objects/parts in different granularity. This hinders further performance optimization of object skeleton detection in complex scenes.

\begin{figure*}[t]
\centering
\includegraphics[width=1.0\textwidth]{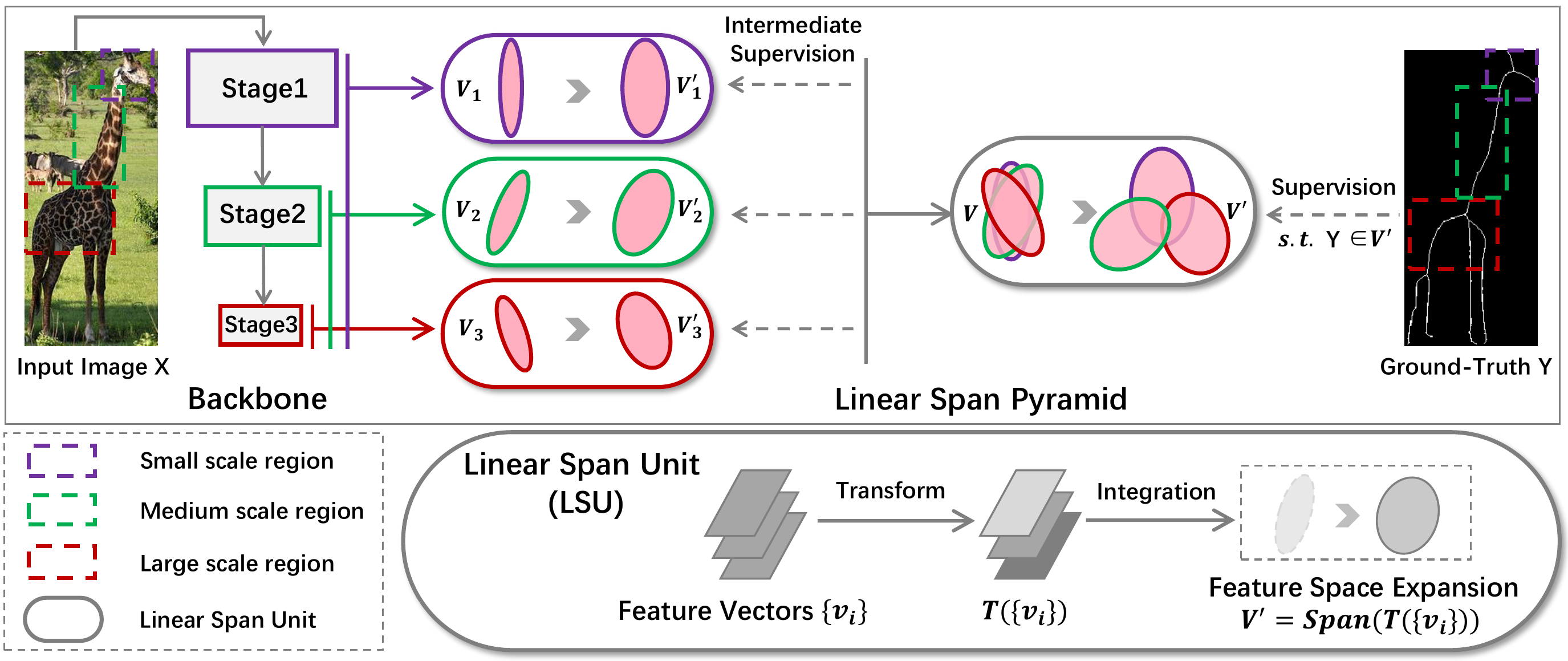}
\caption{Illustration of the architecture search space designed for the proposed adaptive linear span network (AdaLSN). AdaLSN is formulated with the theory of linear span, which aims to perform feature space expansion in both feature subspace (left three color LSUs) and sum-space (right gray LSU) levels. Driven by genetic search algorithm, AdaLSN adaptively configures network architectures and scale-aware features to represent objects in different scale granularity. (Best viewed in color.)}
\label{fig:pipeline}
\end{figure*}

In this paper, we formulate the pixel-wise binary classification tasks as linear reconstruction problems within the linear span framework~\cite{LSN2018} and conclude that the key for network design is to perform feature space expansion. Accordingly, we propose a systematical approach, termed adaptive linear span network (AdaLSN), to transform and integrate hierarchical scale-aware features for object skeleton detection, Fig.~\ref{fig:pipeline}. AdaLSN is driven by neural architecture search (NAS), which optimizes the network operations and connections to push the features towards scale-aware configuration. This improves the feature versatility towards higher accuracy and latency trade-off compared to the state-of-the-arts.

Under the guidance of the linear span theory~\cite{LinearAlgebra2007}, we find out that not only should we expand feature subspace in each stage, but also adaptively reduce structure redundancy to compress their intersection space. Specifically, we design the linear span unit (LSU) to explicitly transform the input features towards complementary with customized operations. Based on LSUs, we construct an explainable search space, termed the linear span pyramid (LSP), to perform subspace and sum-space expansion. LSP is a unit-pyramid mixed search space, which pursues feature complement spaces with a complementary learning strategy. At the unit level, the feature space is expanded by the LSU. At the pyramid level, we progressively pose intermediate supervision to each expanded feature subspace to reduce their semantic overlap and expand the sum-space. With a genetic search algorithm, AdaLSN automatically optimizes the network operations and connections in each area of the search space, including multiple side-outputs, short connections, feature transform, and intermediate supervision, to push the features towards scale-aware configuration. 

Linear Span Network (LSN) was proposed in our previous study~\cite{LSN2018}, while is promoted by introducing a well-designed architecture search space and the adaptive search algorithm. The contributions of this paper are summarized as follows:

\begin{itemize}
    \item We propose the adaptive linear span network (AdaLSN), opening up a promising direction to learn complementary scale-aware features within the framework of linear span.
    
    \item We design a unit-pyramid mixed search space, where competitive network architectures evolve via properly defined genetic operations for search.
    
    \item We improve the state-of-the-arts of object skeleton detection, and demonstrate the general effectiveness of AdaLSN to image-to-mask tasks including edge detection and road extraction.
    
\end{itemize}

%贡献：
%1.结合线性扩张理论，将适用于骨架检测的网络结构设计转化为特征子空间的扩张及其和空间的扩张两个层次。为骨架检测的网络结构设计提供一个明确、有效的设计指导原则。
%2.基于线性扩张理论，设计了cell level 和 side-output level两个层级的搜索空间，并利用遗传算法同时搜索。
%3.搜索得到的网络结构在多个任务和数据集上取得State-of-the-art性能，验证了其通用性

%%%%%%%%%%%%%%%%%%%%%%%%%%%%%%%%%
%%%%%%%%%%%%%%%%%%%%%%%%%%%%%%%%%%

%%%%%%%%%%%%%%%%%%%%%%%%%%%%%%%%%%%%%%%%%%%%%%
% related works
%%%%%%%%%%%%%%%%%%%%%%%%%%%%%%%%%%%%%%%%%%%%%%
\section{Related Work}
Object skeleton detection has attracted much attention in computer vision community for its significance: elaborating object skeleton facilitates image understanding. Existing works typically perform geometric modeling or multi-scale feature fusion to improve the skeleton detection performance, while, to the best of our knowledge, NAS methods have not been considered in this area.

\subsection{Hand-crafted Method}
Early skeleton detection methods were usually performed on binary images by geometric modeling, $e.g.$, morphological image operations. One approach is to treat object skeleton as line subsets which connect super-pixels' center points. Such line subsets were explored from the super-pixels and extract skeleton paths using a sequence of disc models~\cite{Disc2013}. The smoothness of the skeleton can be enforced with spatial filters, $e.g.$, a particle filter, which link local skeleton segments into continuous curves~\cite{ref14}.
When being applied on color images, an image segmentation procedure for contour extraction was first performed as pre-processing. The segmentation procedure tended to produce multi-scale super-pixels, which are converted to skeleton pixels using geometric models.

In the deep learning era, object skeleton detection has been formulated as an pixel-wise binary classification problem with multi-scale feature integration. For clarity, the characters of several typical methods are summarized in Table~\ref{tab.RelatedWork}. The HED~\cite{HED2015} developed parallel multiple side-outputs to produce edges, which can be also used for skeleton detection. Fusing scale-associated deep side-outputs (FSDS)~\cite{FSDS2016} learned skeleton representations with specific scales across multiple network stage given scale-associated ground-truth. 

Side-output residual network (SRN)~\cite{SRN2017} leveraged the side-output residual units to build short connections between adjacent side-output branches for complementary feature utilization. In this way, SRN progressively expand the feature space to fit the errors between the object skeleton/symmetry ground-truth and the multiple side-outputs. To take full use of richer scale-aware features, RCF~\cite{RCF2019}, RSRN~\cite{RSRN2017}, and HiFi~\cite{HiFi2018} establish dense side-output branches for smooth multi-scale feature integration. 

In the linear span view~\cite{LSN2018}, the feature integration actually aims at spanning larger feature spaces for complementary feature extraction and fusion. With this aim, DeepFlux~\cite{DeepFlux2019} utilize the ASPP module~\cite{DeepLabV2} to enrich the semantic level and scale granularity of the feature space. Further, DeepFlux~\cite{DeepFlux2019} contributed a novel strategy by training a network to predict a vector field, which corresponded each image pixel a skeleton/background pixel, in the fashion of flux-based skeletonization. This flux-based skeletonization explicitly encodes the skeletion pixel positions to contextually meaningful entities. Share the same network architecture, GeoSkelNet~\cite{GeoSkeletonNet2019} also used the region-based vector field to model object parts at multiple scales. By designing  Hausdorff distance inspired objective function, both global and local contours were detected through an end-to-end network.

The conventional hand-designed skeleton networks, although incorporating rich prior knowledge, have obvious limitations. With elaborately designed feature integration mechanisms, they are somewhat competent to capture rich representation but still experience difficulty when configuring representation for objects/parts in different scale granularity.

%%%%%%%%%%%%%%%%%%%%%%%%%%%%%%%%%%%%%%%%%%%%%%%%%%%%%%%%%%%%%%%%%%%
%%
%%%%%%%%%%%%%%%%%%%%%%%%%%%%%%%%%%%%%%%%%%%%%%%%%%%%%%%%%%%%%%%%%%%
\begin{table*}[t]
    \centering
    \caption{Summarization of the characters of state-of-the-art approaches for object skeleton detection.}
    \label{tab.RelatedWork}
    \begin{threeparttable}[b]
%    \resizebox{1.0\textwidth}{!}{
    \begin{tabular}{@{}ll|c|llll|llccccc@{}}
    \toprule
          \multicolumn{2}{c|}{\multirow{3}{*}{\textbf{Method}}}
 		& \multirow{3}{*}{\textbf{Year}}
 		& \multicolumn{4}{c|}{\textbf{Network Architecture}}
        & \multirow{3}{*}{\textbf{\tabincell{l}{Multi-scale\\Annotation}}}
        & \multirow{3}{*}{\textbf{\tabincell{l}{Geometric\\Modeling}}}
        \\
        %\hline
        %\cline{4-7}
        &
        &
        & \textbf{\tabincell{l}{Multiple\\Side-outputs}}
        & \textbf{\tabincell{l}{Short\\Connection}}
        & \textbf{\tabincell{l}{Feature\\ Transform}}
        & \textbf{\tabincell{l}{Intermediate\\Supervision}}
\\
        \midrule
		\multirow{9}{*}{\textbf{\tabincell{c}{Manual\\Design}}}
%		 \multirow{9}{*}{\textbf{Manual Designed}}
        & \textbf{HED~\cite{HED2015}}
        & 2015
        & \cmark
        & \xmark
        & \xmark
        & \cmark
        & \xmark
        & \xmark
\\
        
        & \textbf{FSDS~\cite{FSDS2016}}
        & 2016
        & \cmark
        & \cmark
        & \xmark
        & \cmark
        & \cmark
        & \xmark
\\
        
        &  \textbf{RCF~\cite{RCF2019}}
        & 2017
        & \cmark(dense)
        & \xmark
        & \xmark
        & \xmark
        & \xmark
        & \xmark
\\
        
        & \textbf{SRN~\cite{SRN2017}}
        & 2017
        & \cmark
        & \cmark
        & \xmark 
        & \cmark
        & \xmark
        & \xmark
\\
 
        &  \textbf{RSRN~\cite{RSRN2017}}
        & 2017
        & \cmark(dense)
        & \cmark
        & \xmark
        & \cmark
        & \xmark
        & \xmark
\\

%        & \textbf{D-Linknet~\cite{DLinkNet2018}}
%        & 2018
%        & \cmark
%        & \cmark
%        & \cmark 
%        & \cmark
%        & \xmark
%        & \xmark
%\\

        & \textbf{HiFi~\cite{HiFi2018}}
        & 2018
        & \cmark(dense)
        & \cmark
        & \xmark 
        & \xmark
        & \cmark
        & \xmark
\\

        & \textbf{LSN~\cite{LSN2018}}
        & 2018
        & \cmark
        & \cmark(dense)
        & \xmark 
        & \cmark
        & \xmark
        & \xmark

\\
        & \textbf{Deepflux~\cite{DeepFlux2019}}
        & 2019
        & \cmark
        & \xmark
        & \cmark
        & \xmark
        & \xmark
        & Skeleton context flux

\\
        & \textbf{GeoSkelNet~\cite{GeoSkeletonNet2019}}
        & 2019
        & \cmark
        & \xmark
        & \cmark
        & \xmark
        & \xmark
        & Hausdorff distance

\\
        \midrule
         \textbf{\tabincell{c}{Automatic\\Search}}
		& \textbf{AdaLSN (ours)}
        & 2020
        & \cmark(adaptive)
        & \cmark(adaptive)
        & \cmark(adaptive)
        & \cmark(adaptive)
        & \xmark
        & \xmark
\\

    \bottomrule
    \end{tabular}
    \end{threeparttable}
\end{table*}

\subsection{Neural Architecture Search}
NAS targets at optimizing network architectures under the driven of data without human intervene. 
Early NAS methods formulated network designs using reinforcement learning~\cite{zoph2016neural,zoph2018learning,liu2018progressive}, evolutionary algorithms~\cite{real2017large,xie2017genetic}, or random search approaches~\cite{real2018regularized}. %To avoid the inefficiency of encoder-decoder modules, SpineNet~\cite{SpineNet2019} proposed to search a backbone network with scale-permuted intermediate features and cross-scale connections using reinforcement learning. 

To reduce the time cost, one-shot search methods employed weight-reusing~\cite{cai2018efficient} and weight-sharing~\cite{pham2018efficient,Zheng_2019_ICCV} strategies, which fusing the architecture search procedure with network weight optimization. 
A special family of one-shot architecture search, which adjusted itself in continuous spaces, formulated the search space as a super-network~\cite{luo2018neural}. This leads to the differentiate search method where the network and architectural parameters are jointly optimized~\cite{liu2018darts,chen2019progressive,xu2020pcdarts}. 
Recent EfficientNet~\cite{Tan2019efficientnet} defined a scaling method that uniformly searches network depth, width, and resolution by introducing the compound coefficient. 

To automatically fuse multi-scale features, NAS-FPN~\cite{NasFpn2019} defined a search space where the hierarchical side-output features can be optimally combined based on reinforcement learning. NAS-Unet~\cite{NasUnet2019} has the similar idea to fuse hierarchical side-output features by defining primitive operation sets on search space to automatically find cell architectures. For the image segmentation task, Auto-DeepLab~\cite{AutoDeepLab2019} designed a search space including popular hand-designed networks. In the networks, the scale of the convolutional features can be up-sampled, down-sampled or not changed. Driven by gradient-based searching algorithm, Auto-DeepLab found the novel architectures which significantly improved the segmentation performance. 

In this paper, we propose using NAS to define a general feature integration method. By introducing linear span units upon side-output features, we provide the foundation to define operations and connections for feature space expansion. By introducing linear span pyramid, we defines a more complete space for complementary feature extraction and architecture optimization in the linear span framework.
%Within the space, we apply genetic architecture search to jointly optimize unit-level operations and pyramid-level connections in an adaptive fashion. 

%%%%%%%%%%%%%%%%%%%%%%%%%%%%%%%%%%%%%%%%%%%%%%
% method
%%%%%%%%%%%%%%%%%%%%%%%%%%%%%%%%%%%%%%%%%%%%%%

\section{Rethinking Skeleton Network Design}

\subsection{Problem Retrospect}
\label{Sec::PR}
With cascaded convolution and down-sampling operations, CNN backbones typically generate the feature pyramids where features in deeper layers have richer semantics and larger receptive fields, yet fewer fine-details and smaller spatial resolutions. In the pyramid, the semantic levels, scale granularity, and resolution variation of the extracted features tangle together, which raise the following challenges to scale-aware feature integration.
%for object representation.

\begin{itemize}
    \item[1)] How to tackle the large variations of appearance, shape, pose, and scale of objects while depressing cluttered backgrounds with limited convolutional layers?
   
    \item[2)] How to balance the benefit of complementary features and the degradation caused by up-sampling in the resolution alignment for feature integration?

    \item[3)] How to explicitly enrich the semantic hierarchy and scale granularity of the convolutional features?

	\item[4)] How to find out the hierarchical scale-aware features of the largest complementary? 
\end{itemize}

To tackle these challenges, existing methods typically employed the techniques including multiple side-outputs, short connection, feature transform and intermediate supervision, Table ~\ref{tab.RelatedWork}. These techniques have significantly boosted the performance of object skeleton detection. However, the lack of systematic way to integrate these techniques hinders finding out optimal feature representation.

\subsection{Linear Span View}
%介绍骨架检测的基本结构,形式化之后,将其由像素级的二分类问题转化为线性重构问题.
In the deep learning era, object skeleton detection is typically performed with a pre-trained CNN backbone for feature extraction and $1 \times 1$ convolutional layers in the side-output branches for skeleton pixel detection. 

Given an input image $X$, we denote the extracted features as $V_{C \times W \times H} = (v_{c,i,j})_{C \times W \times H}$ and the $1 \times 1$ convolutional layer as $\Lambda_{C \times 1} = (\lambda_{c})_{C \times 1}$, where $C$, $W$, and $H$ respectively represent the channel number, feature map width, and feature map height. Skeleton detection is usually formulated as a pixel-wise classification problem~\cite{HED2015} as
\begin{equation}
    \sum\limits_{c = 1}^{C} {{\lambda_c} \cdot {v_{c,i,j}}} = {\hat {y}}_{i,j} \approx y_{i,j},
   \label{hed::classification}
\end{equation}
where ${\hat {y}}_{i,j}$ and $y_{i,j}$ are respectively the pixel values of the output image ${\hat Y}$ and the ground-truth mask $Y$ at location $(i,j)$. 

Regarding each feature map as a vector $v_c$, where $c$ indexes feature channels, Eq.~\ref{hed::classification} is rewritten from the perspective of linear reconstruction, as 
\begin{equation}
	\sum\limits_{c = 1}^C {{\lambda _c}{v_c}} = \hat{Y} \approx Y.
   \label{hed::reconstruction}
\end{equation}

%承上启下,我们通过两个概念,两个定理的引入,在公式2的基础上建立了我们方法的理论基础,以及搜索空间的设计原则.
Based on Eq.~\ref{hed::reconstruction}, we introduce the linear span concepts and space decomposition theorems~\cite{LinearAlgebra2007}, which lead to the explainable search space and the implementation of AdaLSN.

%定义1讲线性重构问题,转化为特征空间的扩张问题. 继而得出我们需要设计定制的模块来变换扩张集合,以便显式地特征扩张.
\textbf{\textit{Definition 1 (Linear Span Operator).}} \textit{Given a feature vector set $V = \{v_1, v_2, \cdots, v_C\}$ over a field $\mathbb{R}$, the linear span operator is defined as
\begin{equation*}
    \mathcal{V} = Span(V) = \{v|v = \sum\limits_{c=1}^{C} {{\lambda_c}{v_c}}, v_c \in V, \lambda_c \in \mathbb{R} \},
    \label{ls::def1}
\end{equation*}
where set $\mathcal{V}$ constructs a linear space. $V$ denotes the spanning set of $\mathcal{V}$. It is the basis of $\mathcal{V}$, if $V$ is linearly independent.
}

With the \textit{Definition 1}, Eq.~\ref{hed::reconstruction} is updated to
 \begin{equation}
	Span(\{v_1, v_2, \cdots, v_C\}) \ni \hat{Y} \approx Y.
   \label{hed::LinearSpanView}
\end{equation}
Eq.~\ref{hed::LinearSpanView} reveals that each side-output branch of the network can be approximated as a linear system, which is driven by the loss layer to fit the ground-truth.

In the procedure, the key is to optimize the spanning set and implement feature space expansion towards precise ground-truth reconstruction. However, existing approaches, without customized modules to facilitate feature space expansion, are implicit and limited. This inspires us utilizing proper convolutional operators ($O$) to explicitly transform the spanning set while expanding the spanned space, as

\begin{equation}
	Span(O(\{v_1, v_2, \cdots, v_C\})) \ni \hat{Y} \approx Y.
   \label{lsn::transform}
\end{equation}

%LSUs offer customized operations to enhance the advantages of features while suppressing their disadvantages during integration.

%当我们在每个侧输出分支应用设计的模块(LSU)实现了特征空间扩张之后,接下来要考虑的就是如何将他们合并起来. 由空间分解定理,我们将和空间的扩张,转化为浅层的子空间逐步的去拟合深层的子空间的和的补空间,即互补学习策略.
With feature space expansion, we further propose to improve multi-stage feature integration based on the space decomposition and space dimension theorems. 
%complementary learning strategy

%和空间的定义引入,一是空间分解定理用到空间加和的概念,而且空间加和与空间的并不是同一概念,不能不提,虽然定义本身的细节,文中不用.二是特征空间扩张可以在子空间和和空间两个层次上用.
\textbf{\textit{Definition 2 (Sum-space).} } \textit{Suppose $\mathcal{V}_1$ and $\mathcal{V}_2$ are subspaces in the linear space $\mathcal{V}$, the set $\mathcal{V}_{1,2}$, defined as 
\begin{equation*}
    \mathcal{V}_{1,2} = \mathcal{V}_1 + \mathcal{V}_2 = \{v_1 +v_2 | v_1 \in \mathcal{V}_1, v_2 \in \mathcal{V}_2 \},
    \label{ls::def2}
\end{equation*}
is also a linear space, termed the sum-space of $\mathcal{V}_1$ and $\mathcal{V}_2$.}

%%%%%%%%%%%%%%%%%%%%%%%%%%%
%互补学习通过金字塔结构实现了和空间扩张,理论基础就是空间分解定理. 这里是两个空间的例子.多个空间的话,已经监督的深层子空间的和当做一个整体,每次新添加的浅层的子空间就是在拟合其补空间,也就是SRN里面的残差的概念. 定理1说明了补空间(残差)的存在性,以及和空间的基,是子空间的基的并集,也就是说通过跨层短连接就可以合并扩张集合(spanning set) 来实现大空间的扩张分解为小空间的互补扩张再合并.
\textbf{\textit{Theorem 1 (Space Decomposition).}} \textit{Any subspace $\mathcal{V}_{1} \subset \mathcal{V}$ has a complement subspace $\mathcal{V}_{1}^{C}\in\mathcal{V}$ such that 
%$\mathcal{V} = \mathcal{V}_1 + \mathcal{V}_{1}^{C} $ and each vector $v$ in $\mathcal{V}$ can be decomposed uniquely as 
\begin{equation*}
%   v = v_{1} + v_{1}^{c}, v_1 \in \mathcal{V}_{1}, v_{1}^{c} \in \mathcal{V}_{1}^{C}.
\mathcal{V} = \mathcal{V}_1 + \mathcal{V}_{1}^{C},
   \label{ls::theo2}
\end{equation*}	  
and the union of the bases of $\mathcal{V}_{1}$ and $\mathcal{V}_{2}$ is the basis of $\mathcal{V}$. }

Theorem 1 substantiates that we can expand the sum-space of two subspaces by forcing one to fit the complement subspace of the other, which guides the complementary feature learning.
%of multiple network stages. 
%
%Given more subspaces, with the short connections among different side-output branches constructing a division of spanning sets, the intermediate loss layers can assign them different supervision priority. 
%have stronger representation capability while those from the shallow stages are rich in details
%By formulating the features from multiple stages as a sum-space, we propose to progressively 
In specific, we propose to progressively force the feature subspaces from the shallow stages to fit the complement subspace of the deep stages \cite{SRN2017}. Numbering the $S$ stages from shallow to deep, we denote the features extracted from the $s-th$ stage as $V_{s}$, $s = 1, 2, \cdots, S$, the spanned feature subspaces as $\mathcal{V}_{s} = Span(V_{s})$, and the corresponding subspace after transform as $\mathcal{V}_{s}^{'}$, such that $\mathcal{V}_{s}^{'} = Span(O(V_s))$, Eq.~\ref{lsn::transform}. The complementary learning strategy is concluded as  
\begin{equation}
\left\{\begin{array}{l}
\mathcal{V}_{S}^{'} \ni \hat{Y}_S \approx Y\\ 
\mathcal{V}_{S}^{'} + \mathcal{V}_{S-1}^{'} \ni \hat{Y}_{S-1} \approx Y \\
\cdots \\
(\mathcal{V}_{S}^{'} + \mathcal{V}_{S-1}^{'} + \cdots +\mathcal{V}_{2}^{'}) + \mathcal{V}_{1}^{'} \ni \hat{Y}_1 \approx Y 
\end{array}.
\right.
\label{lsn::Complementary Learning}
\end{equation}

%空间维数定理定量的告诉大家,子空间的扩张,及其交空间的压缩都实现了和空间的扩张,一方面从另一个角度证明了之前的LSU和LSP(搭配互补学习策略)这两个设计的合理性.更重要的是,引出了我们要搜索算法去搜的原因,就是LSP太大了,里面有冗余,而且手工调整不可能实现.
%但是要搜什么或者说去掉那些地方的冗余呢,因为这一章去掉了具体实现,所以留到下一章再分析.其实就是之前写的几个adaptive.需要压缩的几个地方,也是可以照应problem retrospect那一节提出的几个挑战的.然后遗传搜索的网络编码就是针对这几个需要压缩的地方来编码去冗余,以实现adaptive.
\textbf{\textit{Theorem 2 (Space Dimension).}} \textit{Supposing $\mathcal{V}$ is a finite dimensional linear space, $\mathcal{V}_1$ and $\mathcal{V}_2$ are two subspaces of $\mathcal{V}$ such that $\mathcal{V} = \mathcal{V}_1 + \mathcal{V}_2 $, and $\mathcal{V}_{1,2}$ is the intersection of $\mathcal{V}_{1}$ and $\mathcal{V}_{2}$, $i.e.$, $\mathcal{V}_{1 \cap 2} = \mathcal{V}_{1} \cap \mathcal{V}_{2} $. 
Then $\mathcal{V}_{1 \cap 2}$ is a linear space, and
\begin{equation*}
   \dim \mathcal{V} = \dim \mathcal{V}_1 + \dim \mathcal{V}_2 - \dim \mathcal{V}_{1 \cap 2}.
\end{equation*}
}

Theorem 2 uncovers that smaller dimension of the intersection space implies bigger dimension of the sum-space. 
This again supports us to perform feature space expansion in the subspace level while compressing their intersection space in the sum-space level. Note that the intersection space of feature subspaces is actually the semantic overlap of multi-stage features, which is mainly caused by architecture redundancy. We propose to utilize architecture encoding and neural architecture search (NAS) to solve these issues.

\begin{figure*}[t]
\centering
\includegraphics[width=.98\textwidth]{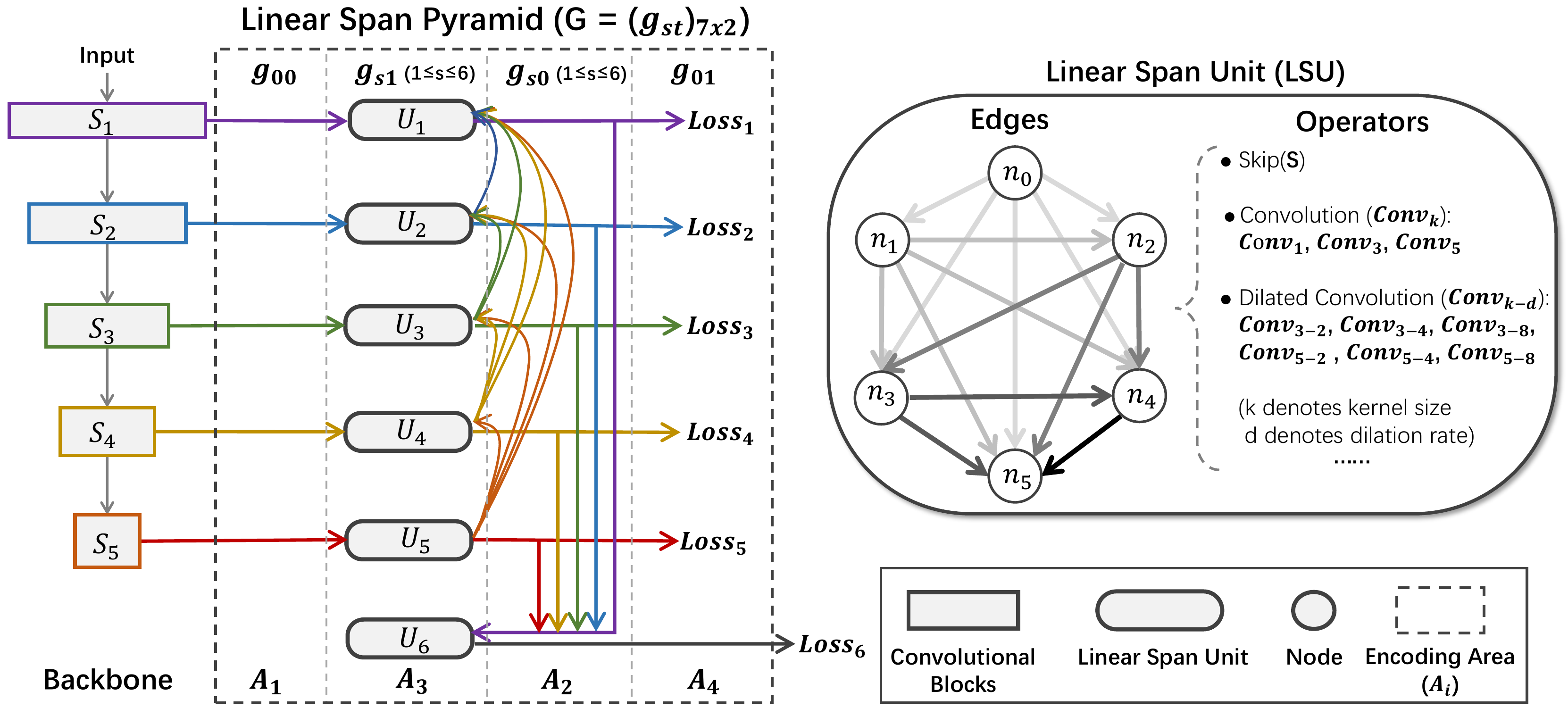}
\caption{Linear span pyramid (LSP) and linear span unit (LSU) constitute the unit-pyramid search space for architecture search and adaptive linear span.
%Red are to be searched: Scale integration, one connection and operator of every two nodes inside four cells.
}
\label{fig:Searchspace}
\end{figure*}

\section{Adaptive Linear Span Network}

Based on the linear span theory~\cite{LinearAlgebra2007}, AdaLSN is materialized by defining a mixed unit-pyramid search space to perform both subspace and sum-space expansion, Fig.~\ref{fig:Searchspace}. To fulfill feature subspace expansion, we design the linear span unit (LSU) which transforms the input feature vectors towards complementary by adaptively selecting proper operators and connections. Based on the LSUs and the complementary learning strategy, we establish the pyramidal architecture search space, which is referred to as the linear span pyramid (LSP). On the pyramid, short connections are utilized to merge the spanning set of each feature subspace towards complementary and complete. 
To reduce the architecture redundancy, we encode the LSP to gene segments for genetic architecture search. The searched AdaLSNs comprehensively integrate the techniques about multiple side-outputs, short connection, feature transform and intermediate supervision with less human experience. 

%%%%%%%%%%%%%%%%%%%%%%%%%%%%%%%%%%%%%%%%%%%%%%%%%%%%%%%
\subsection{Unit-Pyramid Space for Linear Span}

\textbf{Linear Span Unit (LSU).} An LSU can be regarded as a directed acyclic graph, which consists of an input node, an output node, and $P$ intermediate nodes, each of which represents a set of feature vectors, Fig.~\ref{fig:Searchspace}. The nodes are denoted as $n_{i}, i = 0, 1, \cdots, P+1$.
Each pair of nodes has an edge $e_{i_{1}i_{2}}$ connecting the lower-numbered node $n_{i_{1}}$ with the higher-numbered node $n_{i_{2}}$, which consists of an operator $O_{i_{1}i_{2}}$ selected from a pre-defined operator set $\mathcal{O}$. The node $n_{i_{2}}$ is then calculated as $n_{i_{2}} = O_{i_{1}i_{2}}(n_{i_{1}})$. 

To reduce the searching complexity, we set each immediate node to keep only a single edge and the output node $n_{P+1}$ to be the sum of all other $P+1$ nodes. Denote the path from the input node $n_{0}$ to the $i$-$th$ intermediate node as $\mathcal{P}_{i}$ and the node index at $k$-$th$ step in the $i$-$th$ path as $\mathcal{P}_{i}(k)$. Suppose $\mathcal{P}_{i}$ has $K_{i}$ steps, such that $\mathcal{P}_{i}(0) = 0$ and $\mathcal{P}_{i}(K_{i}) = i$. We construct a side-output branch by attaching an LSU $U_{s}$ to the last convolutional layer of the $s-th$ network stage, where the input node $n_{0}^{s}$ and output node $n_{P+1}^{s}$ are actually aforementioned $V_{s}$ and $V_{s}^{'}$ in Eq.~\ref{lsn::Complementary Learning}. Thereafter, the $s$-$th$ LSU can be formulated as
\begin{equation}
\begin{split}
   V_{s}^{'} &= \sum_{i = 0}^{P} (\prod_{k = 1}^{K_{i}}O_{\mathcal{P}_{i}^{s}(k-1),\mathcal{P}_{i}^{s}(k)})(V_{s}) \\
           &= \sum_{i = 0}^{P}O_{\mathcal{P}_{i}^{s}}(V_{s}) \in Span(O^{s}(V_{s})),     
\end{split}
\label{lsu::transform}
\end{equation}
where $O_{\mathcal{P}_{i}^{s}} = \prod_{k = 1}^{K_{i}}O_{\mathcal{P}_{i}^{s}(k-1),\mathcal{P}_{i}^{s}(k)}$ denotes the composite of operators in the path $\mathcal{P}_{i}^{s}$ and $O^{s} = \sum_{i = 0}^{P}O_{\mathcal{P}_{i}^{s}}$ the transformation posed on the input node $n_{0}$, which is defined by the combination of operators at all paths. 

According to Eq.~\ref{lsn::transform}, we conclude that, by selecting proper operators and edges, the unit can transform the input nodes to facilitate feature space expansion. Accommodating the third challenges defined in Sec.~\ref{Sec::PR},  we add skip, convolutions and dilated convolutions with various kernel sizes and dilation coefficients in $\mathcal{O}$ to pertinently enrich the semantic hierarchy and the scale granularity of feature vectors. 
%On the contrary, many existing \empgh{feature transform} modules solely reuse the Atrous Spatial Pyramid Pooling (ASPP)~\cite{DeepLabV2}, which can hardly adapt the difference of features in multiple side-output branches.

\textbf{Linear Span Pyramid (LSP).}
To tackle the challenges in multi-stage feature integration, we materialize the complementary learning strategy by stacking LSUs to construct the pyramid search space with intermediate supervisions.

In specific, according to Eq.~\ref{lsn::Complementary Learning}, we add short connections among LSUs in a deep-to-shallow fashion to merge the spanning sets of the corresponding feature subspaces. Each LSU accepts and accumulates input features from the backbone and/or the output of LSUs from deeper stages, the channels and resolutions of which are respectively aligned to those of the current stage by $1 \times 1$ convolution and upsampling. To ease the noise caused by feature upsampling, we restrict the upsampling rate to $4 \times$ at most, and each LSU only accepts features from two adjacent deeper stages. With intermediate loss layers, LSUs are assigned with different supervision priority to progressively force the feature subspaces to expand towards complementary with each other. We additionally use a fuse LSU ($U_{S+1}$) to accept the outputs from all other LSUs, which further expands the feature sum-space, Fig.~\ref{fig:Searchspace}.

%So far, we can apply cells to pose the first challenge by adaptively configure the combination of operations and edges. Also, in order to tackle the second challenge, we establish a pyramid architecture by stacking cells among stages to perform feature space expansion in both subspace level and sum-space level. As shown in the Fig.~\ref{fig:Searchspace}, we take the commonly used backbone VGG \cite{} to illustrate our pyramid-level search space, where by aligning the resolution of features from different stages, five cells are attached to the last convolution layer of each stage to perform feature subspace expansion, denoted as $C_{i}, i = 1, 2, \cdots, 5$, and an fusion cell is added among all stages to perform feature sum-space expansion, namely $C_{f}$. For feature subspace expansion, each cell $C_{i}$ adaptively integrates feature vectors with higher semantic level and larger scale granularity to enhance the current stage, which expands the subspace $\mathcal{V}_{i}$ to $\mathcal{V}_{i}^{'}$ as

%%%%%%%%%%%%%%%%%%%%%%%%%%%%%%%%%%%%%%%%%%%%%%%%%%%%%%%
\subsection{Adaptive Linear Span by Architecture Search}

Regarding the LSP as a sup-network, Fig.\ \ref{fig:Searchspace}, we aim at instantiating an AdaLSN in the architecture search space to perform linear span. To fulfill this purpose, we propose to encode the LSP into a chromosome matrix and use the genetic algorithm for adaptive architecture search. 

%\textbf{Unit-Pyramid Search Space.}
\textbf{Architecture Encoding.}
LSP consists of a set of operators and connections, which are divided to four encoding areas, Fig.~\ref{fig:Searchspace}. These four encoding areas are respectively defined to handle the four challenges aforementioned in Sec.~\ref{Sec::PR}.

The first encoding area corresponds to connections between the backbone network and the LSUs, which are denoted as a string $g_{00} = \delta_{S_{1}}\delta_{S_{2}} \cdots \delta_{S_{S}}$. 
$\delta$ is a $0$-$1$ binary variable where $\delta_{S_{s}} = 1$ ($1 \leq s \leq S$) means that $U_{s}$ is connected with the $s$-$th$ stage, otherwise to be discarded. The shallow network stages output high resolution features, which require larger memory but have low representation capability, while deep stages output low resolution features with coarse scale granularity, which require smaller memory. Within the first encoding area, we require to adaptively determine connections between the backbone network and the LSUs so that features of shallow and deep stages can be optimally selected.
% but have low representation capability
%众所周知，浅层特征分辨率高，占用显存高，表达能力弱。深层特征分辨率低，占用显存小，但是表达能力强。
%我们定义这个编码区域就是要要自适应搜索stage跟LSU的链接，实现高低分辨率的优化组合。

The second encoding area corresponds to the connections among LSUs, which are represented as a string $g_{s0} = \delta_{s+1}^{s} \cdots \delta_{S+1}^{s}$ and $g_{(S+1)0} = \delta_{U_{1}}\delta_{U_{2}} \cdots \delta_{U_{S}}$. $\delta_{s+1}^{s} \cdots \delta_{S}^{s}$ ($1 \leq s \leq S$) denotes the connection states of unit $U_{s}$ with $U_{s+1}, \cdots, U_{S}$ and $\delta_{U_{s}}$ denotes the connection states of $U_{s}$ with $U_{S+1}$. This encoding area aims to balance the advantage brought by feature complementarity and the degradation caused by the up-sampling operation during multi-stage feature integration. 

The third encoding area contains operators and edges in each LSU to enrich the semantic hierarchy and scale granularity of the feature subspace, which should be adaptive to the characteristics of each stage. This encoding area is denoted as a string $g_{s1} = e_{1}^{s} e_{2}^{s} \cdots e_{P}^{s}o_{1}^{s} o_{2}^{s} \cdots o_{P}^{s}$, $1 \leq s \leq S+1$. $e_{p}$ ($p = 1, 2, \cdots, P$) denotes the index of the node connected to the $p$-$th$ node and $o_{p}$ the corresponding operator on it. $e_{p}$ varies from $0$ to $p-1$ and $o_{p}$ varies from $0$ to $|\mathcal{O}|$.
Based on the above three encoding areas, the LSUs are defined as $U_{s} = (\delta_{S_{s}} \quad g_{s0} \quad g_{s1})$ ($1 \leq s \leq S$) and $U_{S+1} = (g_{(S+1)0} \quad g_{(S+1)1})$.

The fourth encoding area is about the intermediate supervisions, which are denoted as string $g_{01} = \delta_{L_{1}}\delta_{L_{2}} \cdots \delta_{L_{S}}$, where $\delta_{L_{s}}$ ($1 \leq s \leq S$) indicates the connection state of the $U_{s}$ with the $s$-$th$ loss layer. The adaptive architecture search in this encoding area facilitates balancing advantages brought by linear span and the disadvantages ($e.g.$, error accumulation) of intermediate supervisions.
%自适应的决定中间监督能够减少深层特征因上采样造成的这会造成特征degradation.
%为了将深层特征分辨率与输出结果分辨率对应起来，我们需要进行特征上采样，这会造成特征degradation
%第四个编码区域自适应决定哪些中间监督需要保留，
%Considering that deep stages output low resolution features of less fine details, intermediate supervisions bring non-negligible noises to degrade the prediction results that it should be adaptively decided whether or not an LSU should be attached to a loss layer.

%整体编码方案
Combining the four encoding areas, we have the architecture search space for AdaLSN.
The search space is formulated as a matrix ($G = (g_{st})_{(S+2) \times 2}$) with dimensionality $(S+2) \times 2$ where the elements in $G$ are denoted as
\begin{equation}
%\begin{aligned}
%G  &= (g_{ij})_{(S+2) \times 2}, \\
\left\{
\begin{aligned} 
&g_{00} = \delta_{S_{1}}\delta_{S_{2}} \cdots \delta_{S_{S}} , \\
&g_{01} = \delta_{L_{1}}\delta_{L_{2}} \cdots \delta_{L_{S}}, \\
&g_{s0} = \delta_{s+1}^{s} \cdots \delta_{S+1}^{s}, &1 \leq s \leq S,\\
&g_{s1} = e_{1}^{s} e_{2}^{s} \cdots e_{P}^{s}o_{1}^{s} o_{2}^{s} \cdots o_{P}^{s}, &1 \leq s \leq S+1, \\
&g_{(S+1) 0} = \delta_{U_{1}}\delta_{U_{2}} \cdots \delta_{U_{S}},
\end{aligned}
\right.
%\end{aligned}
\label{AdaLSN::encoding}
\end{equation}
where $g_{00}$ corresponds to the multiple side-outputs, $g_{01}$ the intermediate supervision, $g_{s0}$ ($1 \leq s \leq S$) the short connection, and $g_{s1}$ the feature transform. 
The search space covers most side-output network architectures. For example, the well-designed SRN~\cite{SRN2017} falls into this space. SRN contains manual design of multiple side-outputs, short connections between adjacent side-output branches, and intermediate supervisions, which can be represented as
%Fig.~\ref{fig:NetworkRepresentation}, where each LSU has zero intermediate node. 
\begin{equation*}
    G_{SRN} = 
    \left[ \begin{array}{cc}
    11111 & 11111\\
    01000 & - \\
    00100 & - \\
    00010 & - \\
    00001 & - \\
    -     & - \\
    11111 & -
    \end{array}
    \right].
\end{equation*}

\textbf{Architecture Search.}
Architecture search aims to find out sub-structures in the search space for adaptive linear span.
Considering the large dimensionality of the search space, a genetic algorithm is employed for architecture search. In the searching procedure, AdaLSN is defined as a chromosome $G$, the element $g_{ij}$ in the character string matrix $G$ as a gene segment, and each byte in encoded strings as a gene, Eq.~\ref{AdaLSN::encoding}. A chromosome is an individual in the population. Genes and gene segments are the smallest unit for mutation and crossover, respectively. The initial population size is set to 24. Half of the population is randomly selected from all possible chromosomes with equal probability. The other 12 individuals are initialized by mutating the ASPP-like module\cite{DLinkNet2018} once for LSUs and randomly sampling for the rest structures. 

To select superior individuals for evolution, the prediction performance with sufficient training is accurate and reliable but too time-consuming. We resort to sort the loss at one thousand training iterations of the population in each generation and update the 8 individuals with the smallest loss value among all historical generations. The selected top-8 individuals will survive and generate the next generation.

In specific, we apply the crossover and mutation operation on the selected 8 individuals to generate another 8 individuals. To perform crossover, we randomly select a pair of chromosomes and randomly exchange one of their gene segments in the same position. In this way, excellent sub-structures can be probably preserved and evolve to optimal architectures. To avoid local optimum and increase the gene diversity of the population, we randomly change the value of every byte of all gene segments of the chromosomes in the current generation for mutation operation. Every gene in the chromosome has a chance to be changed and excellent gene combinations are compete to be survived and passed to the next generation. In this way, the individuals evolve to adaptively perform feature space expansion for complementary feature learning.

%Every individual in the current generation will be trained for evaluation and selection, then the selected individuals will again to form the next generation iteration. After tens of generations, the search procedure stops and the individual with the smallest training loss outputs as the optimal architecture.
All individuals in the current generation are trained for evaluation and the surviving (selected) individuals will be encoded for the next generation search. After tens of generations, the search procedure stops and the individual with the smallest training loss outputs as the optimal architecture and will be retrained to learn the final model.

\section{Experiments}
In this section, the experimental settings are first described and the modules of AdaLSN are analyzed with ablation studies. The performance of AdaLSN is then presented and compared with the state-of-the-arts. Finally, we apply AdaLSN on edge detection and road extraction to validate its general applicability to image-to-mask tasks.

\begin{table}[t]
    \centering
    \caption{Effect of architecture search. }
    \label{tab.initialization}
    \begin{threeparttable}[b]
    \begin{tabular}{@{}cccc@{}}
    \toprule
        & \textbf{Random Sampling}
        & \multicolumn{2}{c}{\textbf{Architecture Search}}
\\
        &
        & \textbf{Random Init.}
        & \textbf{ASPP-like Init.}
\\
        \midrule
        \textbf{F-score}
        & $0.715\pm0.011$
        & 0.749
        & 0.753\\
    \bottomrule
    \end{tabular}
%    }
    \end{threeparttable}
\end{table}

\begin{table}[t]
    \centering
    \caption{Performance evolution during architecture search.}
    \label{tab.generation}
    \begin{threeparttable}[b]
    \begin{tabular}{@{}cccccccc@{}}
    \toprule
        \textbf{Generation}
        & \textbf{0}
        & \textbf{10}
        & \textbf{20}
        & \textbf{30}
        & \textbf{40}
        & \textbf{50}
        & \textbf{60}
\\
        \midrule
        \textbf{F-score}
        & 0.727
        & 0.744
        & 0.751
        & 0.751
        & 0.750
        & \bf0.753
        & 0.752
\\
    \bottomrule
    \end{tabular}
%    }
    \end{threeparttable}
\end{table}

%%%%%%%%%%%%%%%%%%%%%%%%%%%%%%%%
\begin{comment}
\begin{table}[t]
    \centering
    \caption{Evaluation of encoding areas of AdaLSN.}
    \label{tab.adaptive}
    \begin{threeparttable}[b]
%    \resizebox{1.0\textwidth}{!}{
    \begin{tabular}{@{}lcccc@{}}
    \toprule
         \textbf{}
        & \textbf{$A_{1}$}
        & \textbf{$A_{2}$}
        & \textbf{$A_{3}$}
        & \textbf{$A_{4}$}
%        & \textbf{\tabincell{c}{Complete\\Search}}
        \\
        \midrule
         \textbf{Partial Search (w/o)}
        & 0.727
        & 0.720
        & 0.738
        & 0.750

\\
        \textbf{Random Sampling}
        & 0.739
        & 0.749
        & 0.746
        & 0.750
\\
        \midrule
        \textbf{Complete Search}
        &\multicolumn{4}{c}{75.3}
\\
    \bottomrule
    \end{tabular}
    \end{threeparttable}
\end{table}
\end{comment}

\begin{table}[t]
    \centering
    %at the corresponding area in the network which is searched under the default setting
    \caption{Evaluation of AdaLSN encoding areas. ``Rand." denotes random sampling. Encoding areas \{$A_{1}, A_{2}, A_{3}, A_{4}.$\} are introduced in Fig.~\ref{fig:Searchspace}.}
    \label{tab.adaptive}
    \begin{threeparttable}[b]
    \begin{tabular}{@{}ccccc@{}}
    \toprule

          \textbf{$A_{1}$}
        & \textbf{$A_{2}$}
        & \textbf{$A_{3}$}
        & \textbf{$A_{4}$}
        & \textbf{F-score}
%        & \textbf{\tabincell{c}{Complete\\Search}}
        \\
        \midrule
          \cmark
        & \cmark
        & \cmark
        & \cmark
        & 0.753

\\
        \midrule
        \textbf{Rand.}
        & \cmark
        & \cmark
        & \cmark
        & 0.739
\\
        \textbf{w/o}
        & \cmark
        & \cmark
        & \cmark
        & 0.727
\\
          \midrule
          \cmark
        & \textbf{Rand.}
        & \cmark
        & \cmark
        & 0.749
\\
          \cmark
        & \textbf{w/o}
        & \cmark
        & \cmark
        & 0.720
\\
          \midrule
          \cmark
        & \cmark
        & \textbf{Rand.}
        & \cmark
        & 0.746
\\
          \cmark
        & \cmark
        & \textbf{w/o}
        & \cmark
        & 0.738
\\
          \midrule
          \cmark
        & \cmark
        & \cmark
        & \textbf{Rand.}
        & 0.750
\\
          \cmark
        & \cmark
        & \cmark
        & \textbf{w/o}
        & 0.750
\\
    \bottomrule
    \end{tabular}
    \end{threeparttable}
\end{table}

\begin{figure*}[t]
\centering
\includegraphics[width=\textwidth]{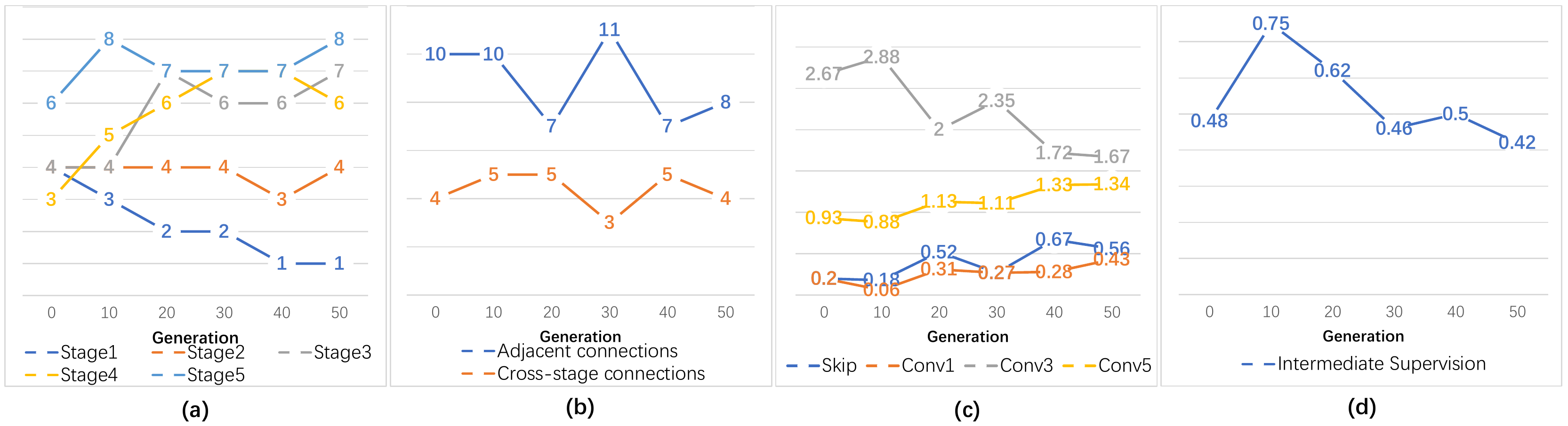}
\caption{Adaptability of architecture search. Each sub-figure shows a statistics of the searched (top-8) architectures. (a) Number of side-outputs attached to each network stage in $A_{1}$; (b) Numbers of short connections in $A_{2}$; (c) Operator distributions (average number of operators in an LSU) in $A_{3}$; (d) Ratio of intermediate supervision in $A_{4}$. }
\label{fig:Adaptability}
\end{figure*}

\subsection{Experimental Setting}

\textbf{Datasets.}
Five commonly used skeleton datasets, including SK-LARGE~\cite{Sklarge2017}, SK-SMALL~\cite{Sksmall2016}, SYM-PASCAL~\cite{SRN2017}, SYMMAX300~\cite{Symmax3002012}, and WH-SYMMAX~\cite{Whsymmax2016}, are used to evaluate AdaLSN. SK-LARGE involves skeletons from about 16 classes of objects, and contains 746/745 training and test images sampled from MS-COCO~\cite{COCO2017}. SK-SMALL (SK506) is sampled from MS-COCO but contains fewer images. SYM-PASCAL contains 648/787 training and test images annotated from the segmentation subset of PASCAL VOC 2011~\cite{Pascal2010}. SYMMAX300 has 200 training images and 100 test images, which are annotated on BSDS300~\cite{BSDS3002001}. WH-SYMMAX is developed for skeleton detection with 228/100 training and test images.

\textbf{Implementation details.}
AdaLSN is implemented using PyTorch and runs on NVIDIA TITAN RTX GPUs (with 24 GB of memory). In the search phase, the raw training set without pre-processing is used to train the architectures in each generation, and fifty valid images are used for evaluation. In both the search and retraining phases, we use the Adam optimizer~\cite{Adam2015} with the initial learning rate 1e-6, a momentum $(0.9, 0.999)$, and a weight decay 0.0005. The batch size is set as 1 while network parameters are updated every 10 times of forward propagation. In the retaining phase, we train the searched architecture for 25 epochs to obtain the final model. The learning rate is fixed during searching and reduced a magnitude after 20 epochs while retraining. We use multiple data augmentation techniques, including resizing to 3 scales (0.8x, 1.0x, and 1.2x), rotating for 4 directions ($0^{\circ}$, $90^{\circ}$, $180^{\circ}$, and $270^{\circ}$), flipping in 2 orientations (left-to-right and up-to-down), and resolution normalization~\cite{GeoSkeletonNet2019}. 

\textbf{Evaluation protocol.}
Following the settings in~\cite{Symmax3002012}, the F-measure score (F-score) and Precision-Recall (PR) curves are used as the evaluation metrics. While PR curves diagnose the binary classification results under different thresholds, F-score provides a single score weighting both precision and recall.

%%%%%%%%%%%%%%%%%%%%%%%%%%%%%%%%%%%%%%%%%%%%%%%%%%%%%
%%%%                  Evolution                  %%%%
%%%%%%%%%%%%%%%%%%%%%%%%%%%%%%%%%%%%%%%%%%%%%%%%%%%%%

\subsection{Ablation Study}
Ablation studies are performed on SK-LARGE~\cite{Sklarge2017}, which is the most used benchmark for object skeleton detection. In the search phase, by default, we set the index of side-outputs as $\{1,2,3,4,5\}$, the channel number of the nodes (Channel Number) as $32$, the intermediate node number (Intermediate Node) as $4$. The kernel size and dilation rate of the convolution layers in the alternative operator set (Operator) vary in $\{1,3,5\}$ and $\{1,2,4,8\}$, respectively.
 
\textbf{Architecture Search.}
%1.搜索算法对初始化是鲁棒的,但ASPP-like更好
In Table~\ref{tab.initialization}, one can see that the searched AdaLSNs significantly outperform the randomly sampled architectures by $\sim3\%$ F-score. AdaLSNs also demonstrate robustness to search initializations. 
%In what following, ASPP-like initialization is used as a default settings. 

In Table~\ref{tab.generation}, we re-train the most excellent individual selected in each population iteration, and report the performance every ten generations. It can be seen that the top-1 individuals evolve quickly and the F-score increases significantly from 0.727 to 0.751, and slightly improves to $0.753$ after 30 more generations. We search fifty generations by default. These results verify the effectiveness of the proposed approach.

%4个编码区的设计是有效的，搜索算法也能收敛。
In Table~\ref{tab.adaptive}, we validate the effectiveness of the encoding areas including multiple side-outputs ($A_{1}$), short connection ($A_{2}$), feature transform ($A_{3}$), and intermediate supervision ($A_{4}$). Specifically, we compare the performance of partial search by screening an encoding area (w/o) and randomly sampling the corresponding area after complete search (Rand.). Experiments show that the encoding areas can boost the performance significantly, validating that the search space is more complete. 

\begin{figure}[t]
\centering
\includegraphics[width=.50\textwidth]{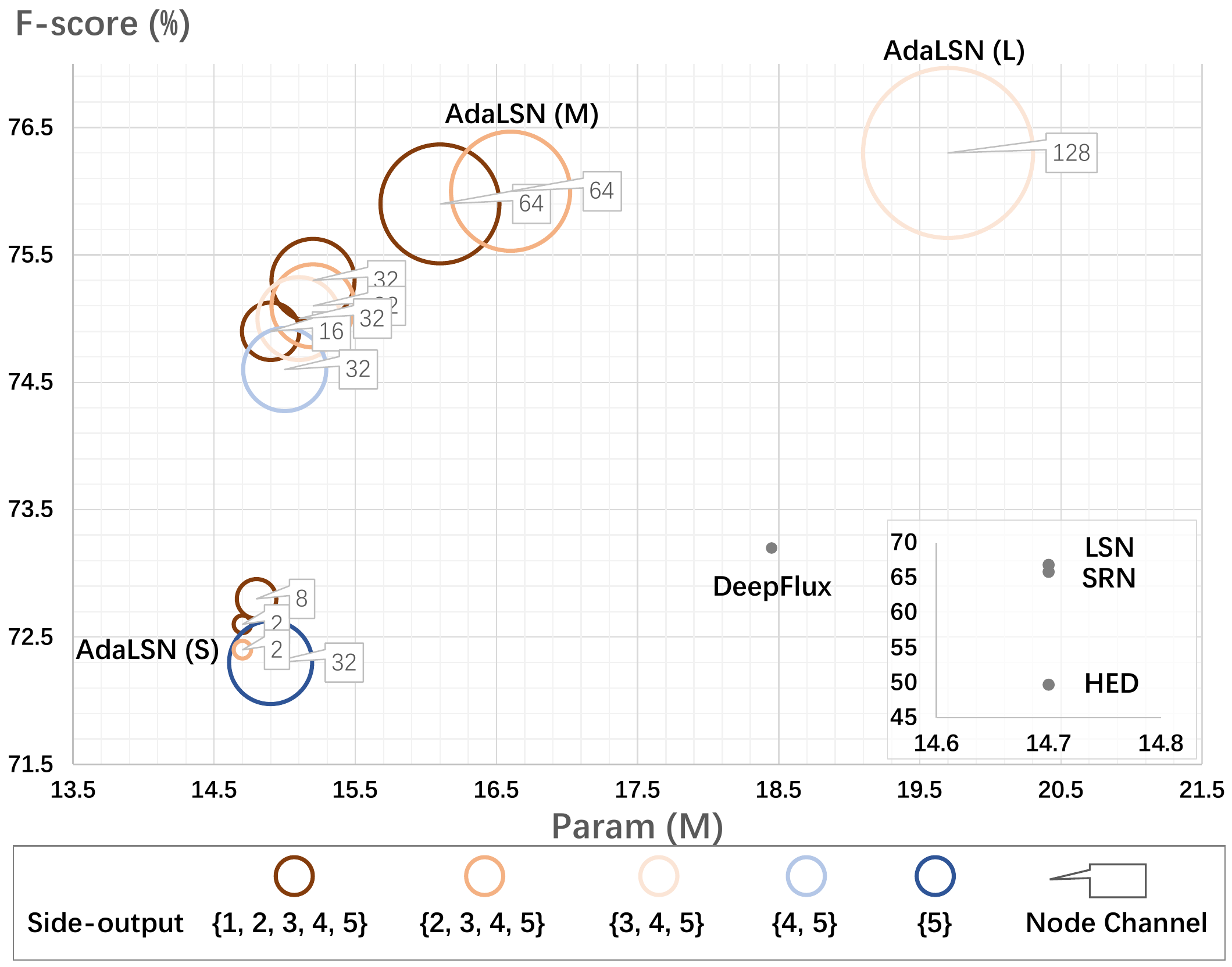}
\caption{Visualization of the parameter and F-score of AdaLSN exemplars with the VGG backbone. The circle size indicates the channel number. (Best view in color and zoom in.)}
\label{fig:variations}
\end{figure}
%%%%%%%%%%%%%%%%%%%%%%%%%%%%%%%%%%%%%%%%%%%%%%%%%%%%%
\begin{table*}[t]
    \centering
    \caption{Ablation study of the proposed AdaLSN with different search space settings.}
    \label{tab.ablation}
    \begin{threeparttable}[b]
    \begin{tabular}{@{}clc|lc|ccc@{}}
    \toprule
         \multicolumn{3}{c}{\textbf{Search Space}}
        & \multicolumn{2}{c}{\textbf{Search Phase}}
        & \multicolumn{3}{c}{\textbf{Retraining Phase}}
        \\
        &
        &
        & \textbf{Memory(G)}
        & \textbf{Search time(h)}
        & \textbf{Param(M)}
        & \textbf{Runtime(ms)}
        & \textbf{F-score}
\\

        \midrule
        \multirow{18}{*}{\textbf{Unit-level}}
        &\multirow{6}{*}{\textbf{Channel Number}}
        & 2
        & $\le 8.1$
        & 43.4
        & 14.7 
        & 12.52
        & 0.726
\\
        &
        & 8
        & $\le 9.9$
        & 46.0
        & 14.8 
        & 12.80
        & 0.728
\\
&
        & 16
        & $\le 12.7$
        & 46.3
        & 14.9
        & 12.56
        & 0.749
\\
&
        & 32
        & $\le 19.8$
        & 47.5
        & 15.2
        & 13.87
        & 0.753
\\
&
        & 64
        & $\approx 24.0$
        & 49.2
        & 16.1
        & 14.21
        & 0.759
\\
&
        & 128
        & $\ge 24.0$
        & --
        & --
        & --
        & --
\\
        %\midrule
        \cline{2-8}
%        \multirow{6}{*}{\textbf{\rotatebox{90}{NODEni}
          %\multirow{6}{*}{\textbf{Unit-level}}
        &\multirow{6}{*}{\textbf{Intermediate Node}}
        \\
        &
        & 0
        & $\le$ 12.8
        & 43.1
        & 14.7
        & 9.13
        & 0.704
\\
&
        & 1
        & $\le$ 13.8
        & 44.0
        & 14.9
        & 9.46
        & 0.740
\\
&
        & 2
        & $\le$ 15.6
        & 45.3
        & 14.9
        & 10.97
        & 0.746
\\
&
        & 3
        & $\le$ 17.3
        & 46.7
        & 15.1
        & 12.03
        & 0.746
\\
&
        & 4
        & $\le$ 19.8
        & 47.5
        & 15.2
        & 13.87
        & 0.753
\\
&
        & 5
        & $\le$ 20.7
        & 49.5
        & 15.2
        & 12.67
        & 0.749
\\
        %\midrule
%        \multirow{5}{*}{\textbf{\rotatebox{90}{OPERATOR}}}
\cline{2-8}
\\
         %\multirow{6}{*}{\textbf{Unit-level}}
        & \multirow{5}{*}{\tabincell{l}{\textbf{Operator} \\\{kernel size\}- \\\{dilation rate\}}}

        & \{1\} - \{1\}
        & $\le 18.8$
        & 45.1
        & 14.8
        & 12.18
        & 0.713
\\
&
        & \{1,3,5\} - \{1\}
        & $\le 19.8$
        & 49.0
        & 15.0
        & 13.32
        & 0.750
\\
&
        & \{1,3,5\} - \{1,2,4,8\}
        & $\le 19.8$
        & 47.5
        & 15.2
        & 13.87
        & 0.753
\\
&
        & \{1,3,5\} - \{1,8,16,24\}
        & $\le 19.8$
        & 48.1
        & 15.1
        & 12.24
        & 0.748
\\
&
        & \{1,3,5\} - \{1,2,4,8,16,24\}
        & $\le 19.8$
        & 46.5
        & 15.0
        & 13.38
        & 0.749
\\
        %\midrule
         \cline{1-8}
         &\\
%       \multirow{5}{*}{\rotatebox{90}{\textbf{STAGE}}}
        \multirow{5}{*}{\textbf{Pyramid-level}}
        & \multirow{5}{*}{\textbf{Side-output}}
        & \{1,2,3,4,5\}
        & $\le 19.8$
        & 47.5
        & 15.2
        & 13.87
        & 0.753
\\
&
        & \{2,3,4,5\}
        & $\le 13.2$
        & 42.2
        & 15.2
        & 14.42
        & 0.751
\\
&
        & \{3,4,5\}
        & $\le 11.3$
        & 39.5
        & 15.1
        & 12.34
        & 0.750
\\
&
        & \{4,5\}
        & $\le 10.3$
        & 38.4
        & 15.0
        & 12.13
        & 0.746
\\
&
        & \{5\}
        & $\le 9.8$
        & 37.5
        & 14.9
        & 10.12
        & 0.723
\\

%%%%%%%%%%%%%%%%%%%%%%%%%%%%%%%%%%%%%%%%%%%%
%% stages
%%%%%%%%%%%%%%%%%%%%%%%%%%%%%%%%%%%%%%%%%%%%
    \bottomrule
    \end{tabular}
    \end{threeparttable}
\end{table*}
%%%%%%%%%%%%%%%%%%%%%%%%%%%%%%%%%%%%%%%%%%%%%%%%%

%4.得到AdaLSN的三个版本(按搜索空间的大中小分)
%%%%%%%%%%%%%%%%%%%%%%%%%%%%%%%%%%%%%%%%%%%%%%
%%%%             Pareto curve             %%%%
%%%%%%%%%%%%%%%%%%%%%%%%%%%%%%%%%%%%%%%%%%%%%%

%%%%%%%%%%%%%%%%%%%%%%%%%%%%%%%%%%%%%%%%%%%%%%%%%
\begin{table*}[t]
    \centering
    \caption{Comparison of AdaLSN exemplars with different search space settings and backbone networks.}
    \label{tab.variations}
    \begin{threeparttable}[b]
    \resizebox{1.0\textwidth}{!}{
    \begin{tabular}{@{}clccc|lc|ccc@{}}
    \toprule
        &\multicolumn{4}{c}{\textbf{Search Space}}
        & \multicolumn{2}{c}{\textbf{Search Phase}}
        & \multicolumn{3}{c}{\textbf{Retraining Phase}}
\\

        & \textbf{Backbone}
        & \textbf{Side-output}
        & \textbf{Channel Number}
        & \textbf{Intermediate Node}
        & \textbf{Memory(G)}
        & \textbf{Search time(h)}
        & \textbf{Param(M)}
        & \textbf{Runtime(ms)}
        & \textbf{F-score}
\\
        \midrule
         (S)
        &  \textbf{VGG16}
        & \{2,3,4,5\}
        & 2
        & 1
        & $\le 7.2$
        & 43.1
        & 14.7
        & 12.22
        & 0.724
\\
         (M)
        &  \textbf{VGG16}
        & \{2,3,4,5\}
        & 64
        & 4
        & $\le 16.0$
        & 45.3
        & 16.6
        & 14.67
        & 0.760
\\
         (L)
        &  \textbf{VGG16}
        & \{3,4,5\}
        & 128
        & 4
        & $\le 19.2$
        & 47.2
        & 19.7
        & 15.53
        & 0.763
\\
%%%%%%%%%%%%%%%%%%%%%%%%%%%%%%%%%%%%%%%%%%%%
%% Vgg
%%%%%%%%%%%%%%%%%%%%%%%%%%%%%%%%%%%%%%%%%%%%
%        \midrule
         (L)
        &\multirow{1}{*}{\textbf{ResNet50}}
        & \{3,4,5\}
        & 128
        & 4
        & $\approx 24.0$
        & 138.7
        & 30.9
        & 43.12
        & 0.764
\\
%%%%%%%%%%%%%%%%%%%%%%%%%%%%%%%%%%%%%%%%%%%%
%% resnet50
%%%%%%%%%%%%%%%%%%%%%%%%%%%%%%%%%%%%%%%%%%%%
%        \midrule
         (L)
        &\multirow{1}{*}{\textbf{Res2Net}}
        & \{3,4,5\}
        & 128
        & 4
        & $\approx 24.0$
        & 117.3
        & 26.1
        & 34.23
        & 0.768
\\
%%%%%%%%%%%%%%%%%%%%%%%%%%%%%%%%%%%%%%%%%%%%
%% Res2Net
%%%%%%%%%%%%%%%%%%%%%%%%%%%%%%%%%%%%%%%%%%%%
%        \midrule
         (L)
        &\multirow{1}{*}{\textbf{InceptionV3}}
        & \{3,4,5\}
        & 128
        & 4
        & $\approx 24.0$
        & 193.2
        & 32.4 
        & 69.37
        & 0.786
\\
%%%%%%%%%%%%%%%%%%%%%%%%%%%%%%%%%%%%%%%%%%%%
%% inceptionV3
%%%%%%%%%%%%%%%%%%%%%%%%%%%%%%%%%%%%%%%%%%%%
%        \midrule
%        \multirow{1}{*}{\textbf{ResNeSt}}
%        & \{3-5\}-\{128\}-\{4\}
%        & $\approx 24.0$
%        & 
%        & 32.8
%        & 0.03913
%        & 0.738
%\\
%%%%%%%%%%%%%%%%%%%%%%%%%%%%%%%%%%%%%%%%%%%%
%% ResNest
%%%%%%%%%%%%%%%%%%%%%%%%%%%%%%%%%%%%%%%%%%%%

    \bottomrule
    \end{tabular}
    }
    \end{threeparttable}
\end{table*}

%%%%%%%%%%%%%%%%%%%%%%%%%%%%%%%%%%%%%%%%%%%%%%%%%

\textbf{Architecture Adaptability.}
In Fig.~\ref{fig:Adaptability}, we validate the adaptability of the four encoding areas using statistics figures. In $A_{1}$, Fig.~\ref{fig:Adaptability}a shows that the side-output numbers in deep stages are significantly larger than those in shallow stages. This is because the features from deep stages have small resolutions yet coarse-scale granularity and strong representation capability. More deep stage features imply higher accuracy. The adaptive configuration of multi-stage features greatly eases the scale variation problem. In $A_{2}$, we divide the short connections adjacent ones and cross-stage ones, Fig.~\ref{fig:Adaptability}b. Considering the degradation caused by feature upsampling, the adjacent connections are preferred.

In $A_{3}$, we compare LSU operator distributions, including skips and convolutions with different kernel sizes, Fig.~\ref{fig:Adaptability}c. For the moderate performance gain (0.750 $vs$ 0.753) with dilated convolutions, we calculate the average number of different convolutions in an LSU according to their kernel size only. It can be seen that the $Conv_{3}$ and $Conv_{5}$ are more preferred than $Conv_{1}$. This is because they correspond to stronger representation capability and larger receptive fields to ease the imbalance of semantic level and scale granularity. In the early search phrase, the number of $Conv_{3}$ is noticeably larger than that of $Conv_{5}$ while the gap reduces when search goes on. The reason lies in that with less parameters, $Conv_{3}$ layers are easy to be optimized. When sufficient evolution takes place, $Conv_{5}$ with large receptive field and representation capability show its advantages. This validates AdaLSN's adaptability on network architecture configuration. 

In $A_{4}$, we calculate the ratio of intermediate supervision, Fig.~\ref{fig:Adaptability}d. It can be seen that in the early phase, the architecture is more dependent on intermediate supervision because intermediate supervision can force the feature subspaces to expand to fit the ground-truth. With sufficient training after tens of generations, architectures with less intermediate supervision can ease the error accumulation issue while fulfilling complementary learning for feature space expansion.

%%%%%%%%%%%%%%%%%%%%%%%%%%%%%%%%%%%%%%%%%%%%%%
%%%%        visualization of SML          %%%%
%%%%%%%%%%%%%%%%%%%%%%%%%%%%%%%%%%%%%%%%%%%%%%
\begin{figure*}[t]
\centering
\includegraphics[width=\textwidth]{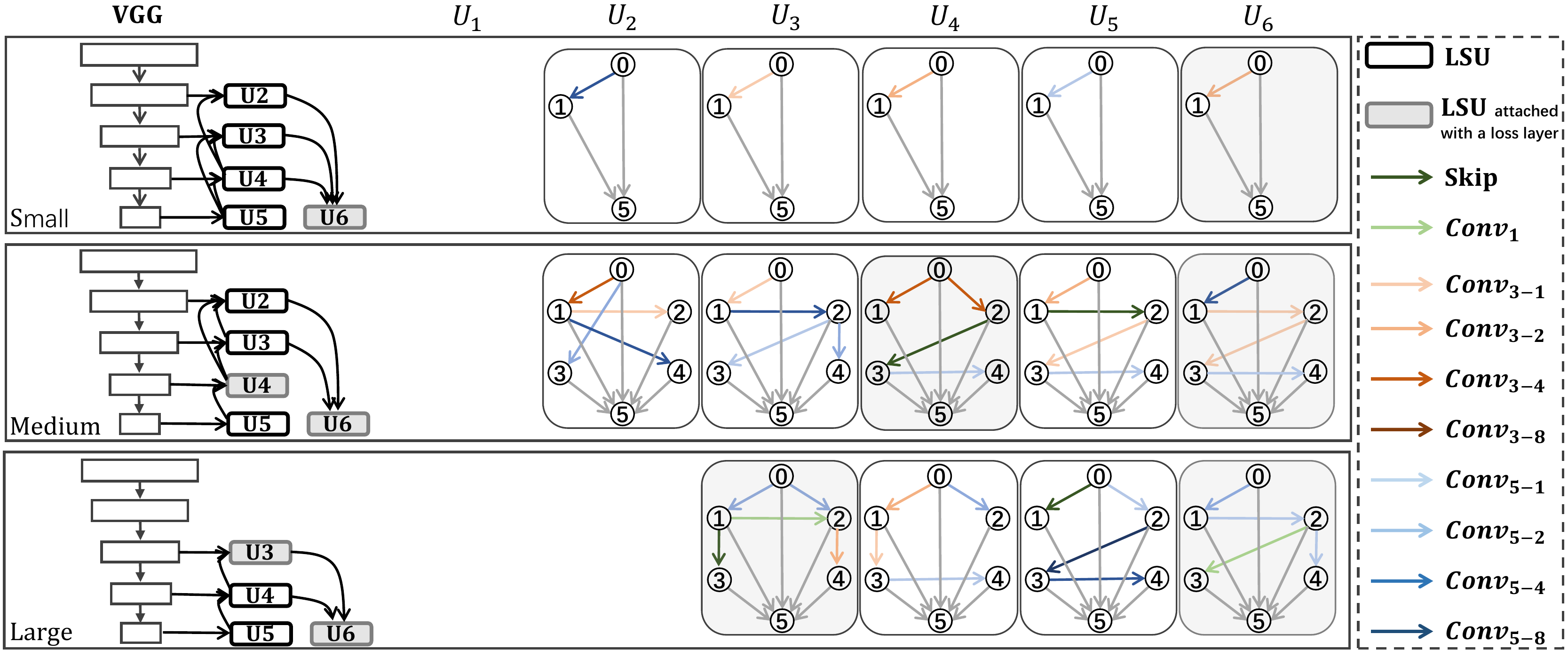}
\caption{The searched exemplar architectures of AdaLSN.}
\label{fig:AdaLSNexampler}
\end{figure*}

\textbf{AdaLSN Exemplars.}
The computational cost and the prediction accuracy of AdaLSNs largely depend on the search space settings, Table~\ref{tab.ablation}. Specifically, we explore the effect of ``channel number", ``intermediate node", and ``operator" (\{kernel size\}-\{dilation rate\}) at the unit level; and the effect of ``side-output" at the pyramid level. 

When the channel number increases from 2 to 64, the memory cost increases from $8.1G$ to $24.0G$. The parameters and runtime of the searched network in the retraining phase increase moderately from ($14.7M$, $12.52ms$) to ($16.1M$, $14.21ms$), while the F-score increases significantly from $0.726$ to $0.759$, Table~\ref{tab.ablation}. We conclude the channel number is an important factor in balancing the computational cost and accuracy for our approach.

When adding intermediate nodes in each LSU from $0$ to $4$, one can see that the computational cost in both phases increases moderately, while the F-score greatly increases from $0.704$ to $0.753$. It is worth noting that comparing LSUs without any intermediate node, the F-score increases impressively from $0.704$ to $0.740$ with only $1$ intermediate node used in each LSU. This validates the importance of operators and their combinations in the explicit feature transform for feature space expansion. However, when the intermediate node number increases to $5$, the F-score of AdaLSN falls to $0.749$. This is because the more intermediate nodes means higher search complexity, which significantly improves the difficult to search optimal architectures. 

We evaluate the convolution operators with different kernel sizes and dilation rates. Table~\ref{tab.ablation} shows that convolutions with kernel sizes larger than $1$ have superiority for feature space expansion, $e.g.$, the F-score increases from $0.713$ to $0.750$ with negligible computational cost. When the dilation rate enlarges properly as $\{1, 2, 4, 8\}$, which enriches the scale granularity, the performance improves up to $0.753$. Nevertheless, the F-score falls upon larger dilation rates, because convolution operators with large dilation tend to missing feature details. Accordingly, the kernel sizes and dilation rates are set as $\{1, 3, 5\}-\{1, 2, 4, 8\}$.

As features from shallow stages are less representative but have high resolution, we gradually reduce the shallow side-outputs in the search phase to figure out the effect of each side-output. As shown in the bottom row of Table~\ref{tab.ablation}, without $S_{1}$ and $S_{2}$, the memory cost reduces significantly from $19.8G$ to $11.3G$ and the search time decreases from $47.5h$ to $39.5h$, while the F-score reduces slightly. 
In Fig.~\ref{fig:variations}, we visualize the figures of ablation on channel number and side-output in Table~\ref{tab.ablation} and Table~\ref{tab.variations} using the VGG backbone. One can find that when reducing the shallow network stages $\{1, 2\}$ while increasing the channel number from $32$ to $128$, the F-scores of the searched architectures increase significantly. When we reduce the deep stages with medium channel numbers, the F-score significantly drops. For instance, with the channel number $32$, the F-score drops from $0.746$ to $0.723$ when the side-outputs reduce from $\{4, 5\}$ to $\{5\}$. 

Empirically, we set the side-output as $\{2, 3, 4, 5\}$ and change the channel number and intermediate node to (2, 1) for a small AdaLSN($S$) , while (64, 4) for a medium AdaLSN($M$). By increasing the channel number to $128$ while reducing the side-output to $\{3, 4, 5\}$, we have a large AdaLSN($L$).  The detailed structures of three versions of AdaLSN with VGG~\cite{VGG2015} are depicted in Fig.~\ref{fig:AdaLSNexampler}. Comparing to the state-of-the-art approaches, such as HED~\cite{HED2015}, SRN~\cite{SRN2017}, and LSN~\cite{LSN2018}, AdaLSN($S$) has significant accuracy improvement (about $0.056\sim0.227$ F-score gain) with negligible parameter cost. Comparing with DeepFlux~\cite{DeepFlux2019}, AdaLSN($M$) achieves better F-score (0.760 $vs$ 0.724) with less params (16.6 M $vs$ 18.3 M) . In Table~\ref{tab.variations}, we can find that with the InceptionV3 ~\cite{inceptionV32016} backbone, AdaLSN($L$) achieves 0.786, improving the state-of-the-art with a large margin.
%%%%%%%%%%%%%%%%%%%%%%%%%%%%%%%%%%%%%%%%%%%%%%%%%%%%%

%%%%%%%%%%%%%%%%%%%%%%%%%%%%%%%%%%%%%%%%%%%%%%%%%%%%%

%%%%%%%%%%%%%%%%%%%%%%%%%%%%%%%%%%%%%%%%%%%
\begin{table*}[t]
%\begin{threeparttable}
    \centering
    \caption{Performance comparison of state-of-the-art approaches on commonly used skeleton$\backslash$symmetry detection datasets.}
    \label{tab.sota}
    \begin{threeparttable}[b]
%    \resizebox{\textwidth}{!}{
    \begin{tabular}{@{}lccccc@{}}
    \toprule
         \multirow{2}{*}{\textbf{Methods}}
        & \multicolumn{5}{c}{\textbf{Datasets}} \\

        &\textbf{SK-LARGE~\cite{Sklarge2017}}
        & \textbf{SK-SMALL~\cite{Sksmall2016}}
        & \textbf{WH-SYMMAX~\cite{Whsymmax2016}}
        & \textbf{SYM-PASCAL~\cite{SRN2017}}
        & \textbf{SYMMAX300~\cite{Symmax3002012}}

        \\
        \midrule
       \textbf{MIL~\cite{MIL2012}}

        &  0.353
        &  0.392
        &  0.365
        &  0.174
        &  0.362\\
        
        \textbf{HED~\cite{HED2015}}

        &  0.497
        &  0.541
        &  0.732
        &  0.369
        &  0.427\\
        
        \textbf{RCF~\cite{RCF2019}}

        &  0.626
        &  0.613
        &  0.751
        &  0.392
        &  -\\
        
       \textbf{FSDS~\cite{FSDS2016}}

        &  0.633
        &  0.623
        &  0.769 
        &  0.418
        &  0.467\\
        
        \textbf{SRN~\cite{SRN2017}}

        &  0.658
        &  0.632
        &  0.780
        &  0.443
        &  0.446\\
        
        \textbf{LSN~\cite{LSN2018}}

        &  0.668
        &  0.633
        &  0.797
        &  0.425
        &  0.480\\
        
       \textbf{Hi-Fi~\cite{HiFi2018}}

        &  0.724
        &  0.681
        &  0.805
        &  0.454
        &  -\\
        
        \textbf{DeepFlux~\cite{DeepFlux2019}}

        &  0.732
        &  0.695 
        &  0.840
        &  \bf0.502
        &  0.491\\
        
 %       \textbf{GeoSkelNet~\cite{GeoSkeletonNet2019}}

 %       &  0.757
 %       &  0.727
 %       &  0.849
 %       &  0.520
 %       &  0.501\\

        \midrule
        
        \textbf{AdaLSN (ours)}

        &  \bf0.786
        &  \bf0.740
        &  \bf0.851
        &  0.497
        &  \bf0.495\\
    \bottomrule
    \end{tabular}
%    }
    \end{threeparttable}
\end{table*}
%%%%%%%%%%%%%%%%%%%%%%%%%%%%%%%%%%%%%%%%%%%
%%%%%%%%%%%%%%%%%%%%%%%%%%%%%%%%%%%%%%%%%%%%%%
%%%%         linear span            %%%%
%%%%%%%%%%%%%%%%%%%%%%%%%%%%%%%%%%%%%%%%%%%%%%

\begin{figure}[t]
\centering
\includegraphics[width=0.48\textwidth]{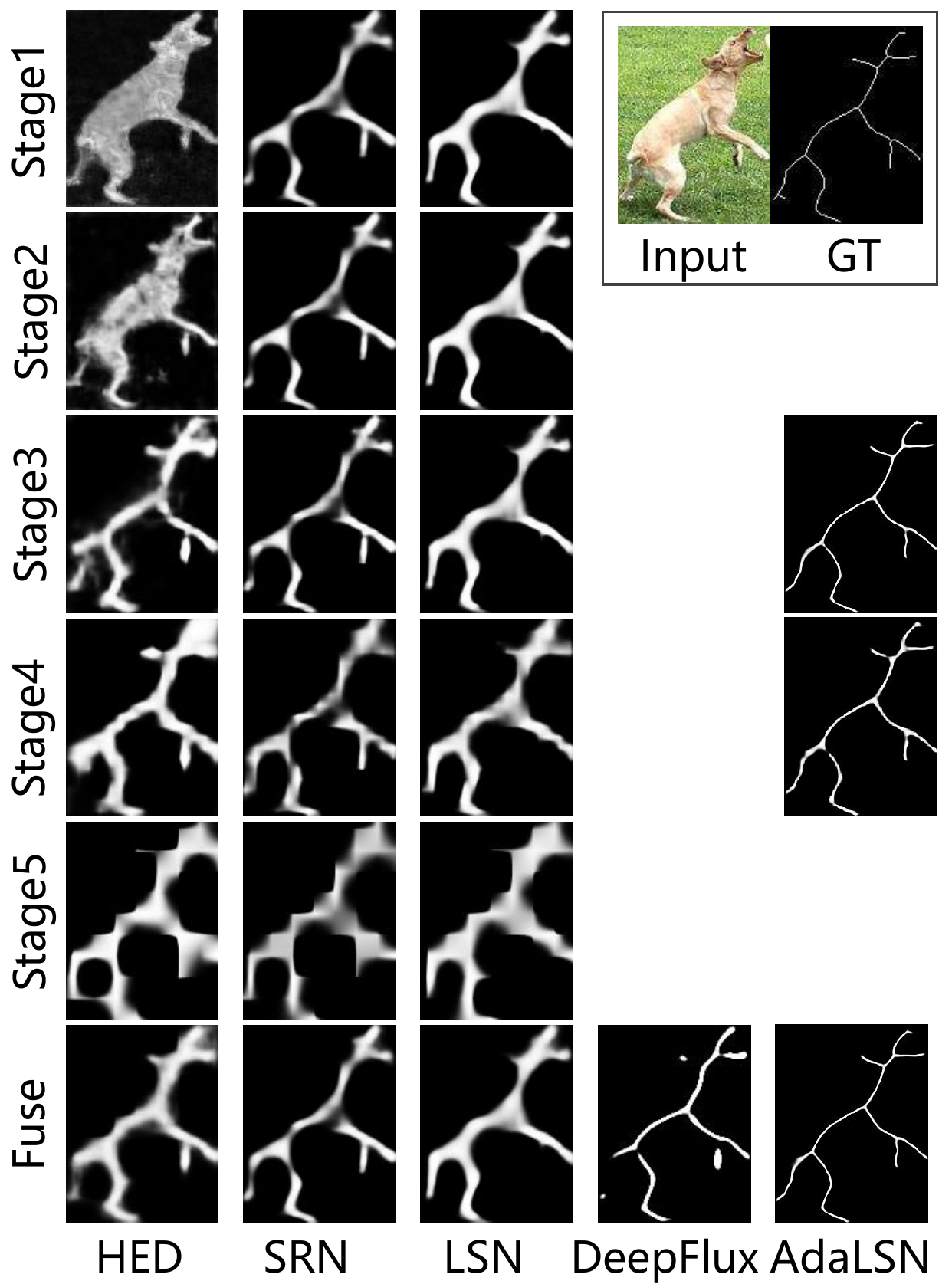}
\caption{Comparison of predicted masks, which reflect the feature space expansion in different feature stages.}
\label{fig:linearspaneffect}
\end{figure}
%%%%%%%%%%%%%%%%%%%%%%%%%%%%%%%%%%%%%%%%%%%%%%%%%%%%%
\begin{figure*}[t]
\centering
\includegraphics[width=\textwidth]{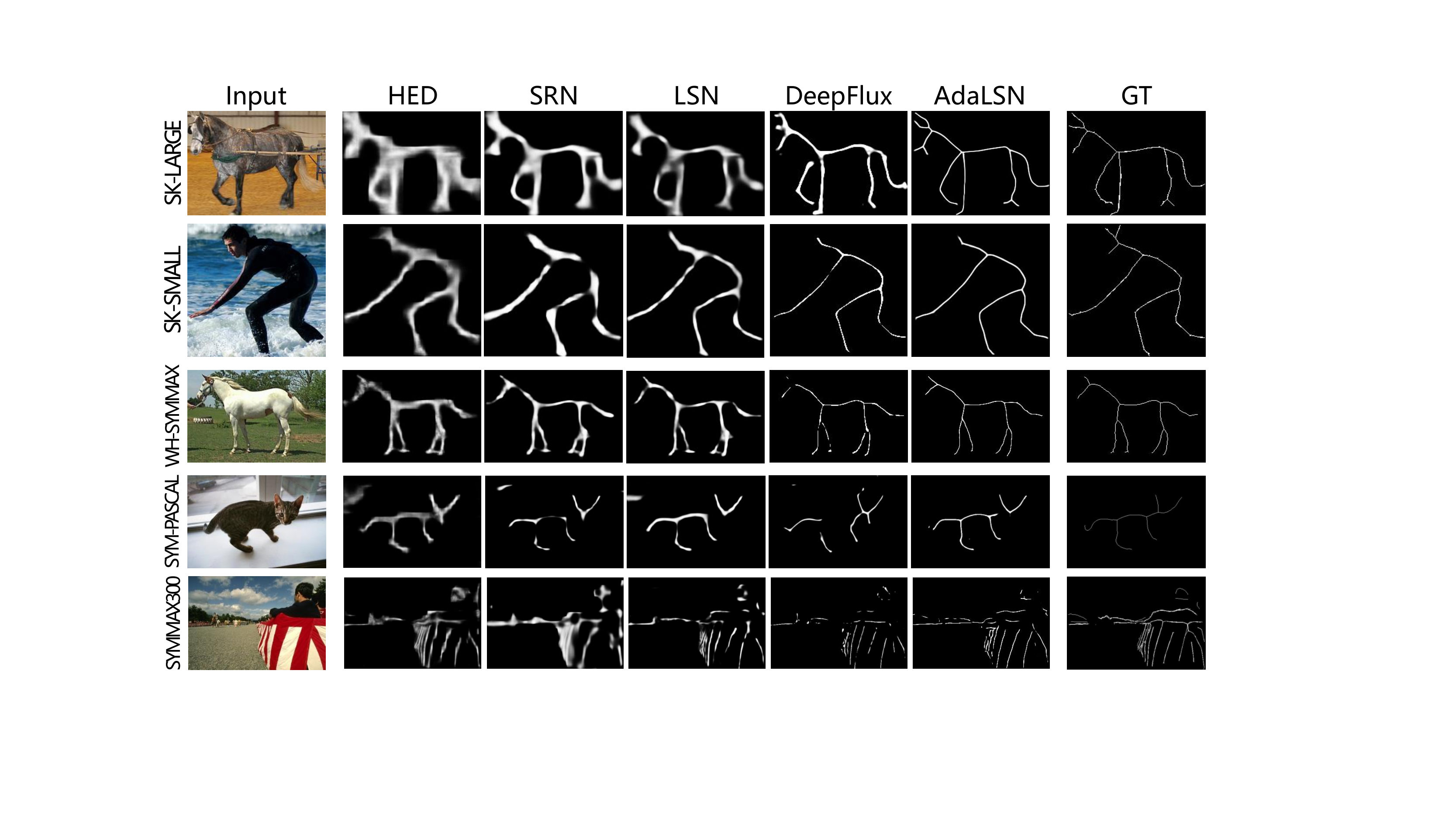}
\caption{Skeleton detection examples by state-of-the-art approaches on commonly used skeleton detection datasets including SK-LARGE~\cite{Sklarge2017}, SK-SMALL~\cite{Sksmall2016}, WH-SYMMAX~\cite{Whsymmax2016}, SYM-PASCAL~\cite{SRN2017}, and SYMMAX300~\cite{Symmax3002012}.}
\label{fig:SkeletonResult}
%\vspace{-0.2cm}
\end{figure*}

\textbf{Linear Span.}
In Fig.~\ref{fig:linearspaneffect}, we compare the skeleton predictions of the state-of-the-art approaches about a dog. It can be seen that with only parallel side-outputs for scale-aware feature utilization, the predictions of HED~\cite{HED2015} suffer background noise in shallow network stages and mosaic effects in deep stages. With adding short connections among side-outputs for feature integration, SRN~\cite{SRN2017} purses the residual between adjacent stages to progressively improve the predictions in a deep-to-shallow manner so that the predictions in shallow stages are greatly improved. To learn more complementary features and span a larger feature space, LSN~\cite{LSN2018} builds dense short connections among side-outputs for feature space expansion. However, without explicitly feature transform, LSN only moderately improves the result comparing with SRN. Deepflux~\cite{DeepFlux2019} generates a slim skeleton with manually designed ASPP modules~\cite{DeepLabV2} and the context flux constrain. However, the prediction is also barely satisfactory.

AdaLSN incorporates these advantages in the linear span view and updates them with the adaptive architecture search algorithm. As the searched architectures are adaptive to the skeleton characteristics, feature subspaces are adaptively expanded while forced to be complementary with each other. This explains why the prediction results at all stages are more precise, clear, and consecutive.
%%%%%%%%%%%%%%%%%%%%%%%%%%%%%%%%%%%%%

\subsection{Performance and Comparison}

The proposed AdaLSN approach outperforms the CNN based state-of-the-art methods in terms of F-score, Tab.~\ref{tab.sota}. The results of AdaLSN is reported by the architecture searched in the SKLARGE with the large version setting of the inceptionV3~\cite{DeepLabV2} backbone and retrained on each dataset.

With the SKLARGE dataset, one can find that Hi-Fi~\cite{HiFi2018} with additional scale-associated ground-truth achieves the F-score of 72.4\%; DeepFlux~\cite{DeepFlux2019} with skeleton context flux reports the highest skeleton detection performance up to date of 73.2\%. Without additional supervision information and explicitly geometric modeling, AdaLSN achieves greatly high F-score of 78.6\%, which outperforms DeepFlux by significant margins of 5.4\%. When comparing to DeepFlux with transferring the architecture to other skeleton$\backslash$symmetry datasets, AdaLSN respectively improves the F-score by 4.5\%, 1.1\% and 0.4\% on SK-SMALL, WH-SYMMAX, and SYMMAX300, and achieves comparable performance on SYM-PASCAL. 

The detected skeleton results are shown and compared in Fig~\ref{fig:SkeletonResult}, where our method can extract the skeleton maps of different granularity with better continuity and higher accuracy. In specific, HED produces skeletons with lots of noise. SRN and LSN predict relatively clearer skeletons which are not smooth. Deepflux reports clearer and slimmer results, while remains some false positive points and dis-continual segments. Incorporating complementary features rich in semantic hierarchy and scale granularity, AdaLSN produces more precise skeleton masks similar to the groundtruth.

%%%%%%%%%%%%%%%%%%%%%%%%%%%%%%%%%%%%%%%%%%%%%%%%%
%%                 Other Tasks                 %%
%%%%%%%%%%%%%%%%%%%%%%%%%%%%%%%%%%%%%%%%%%%%%%%%%
\subsection{Other Image-to-Mask Tasks}
AdaLSN can be directly applied in other image-to-mask task such as edge detection and road extraction. In this section, we compare the edge detection and road extraction performance of the proposed AdaLSN with some other state-of-the-art methods. The good performance of the proposed AdaLSN in these two tasks demonstrates its general applicability to image-to-mask tasks.% such as Canny\cite{Canny1986}, Sketech Tokens [10], Structured Edge (SE) [3], gPb [1], DeepContour\cite{DeepContour2015}, HED\cite{HED2015}, and SRN\cite{SRN2017}, Fig. 8 and Table 5. 

%%%%%%%%%%%%%%%%%%%%%%%%%%%%%%%%%%%%%%%%%%%%%%%%%%%%%
\begin{figure}[t]
\centering
\includegraphics[width=.5\textwidth]{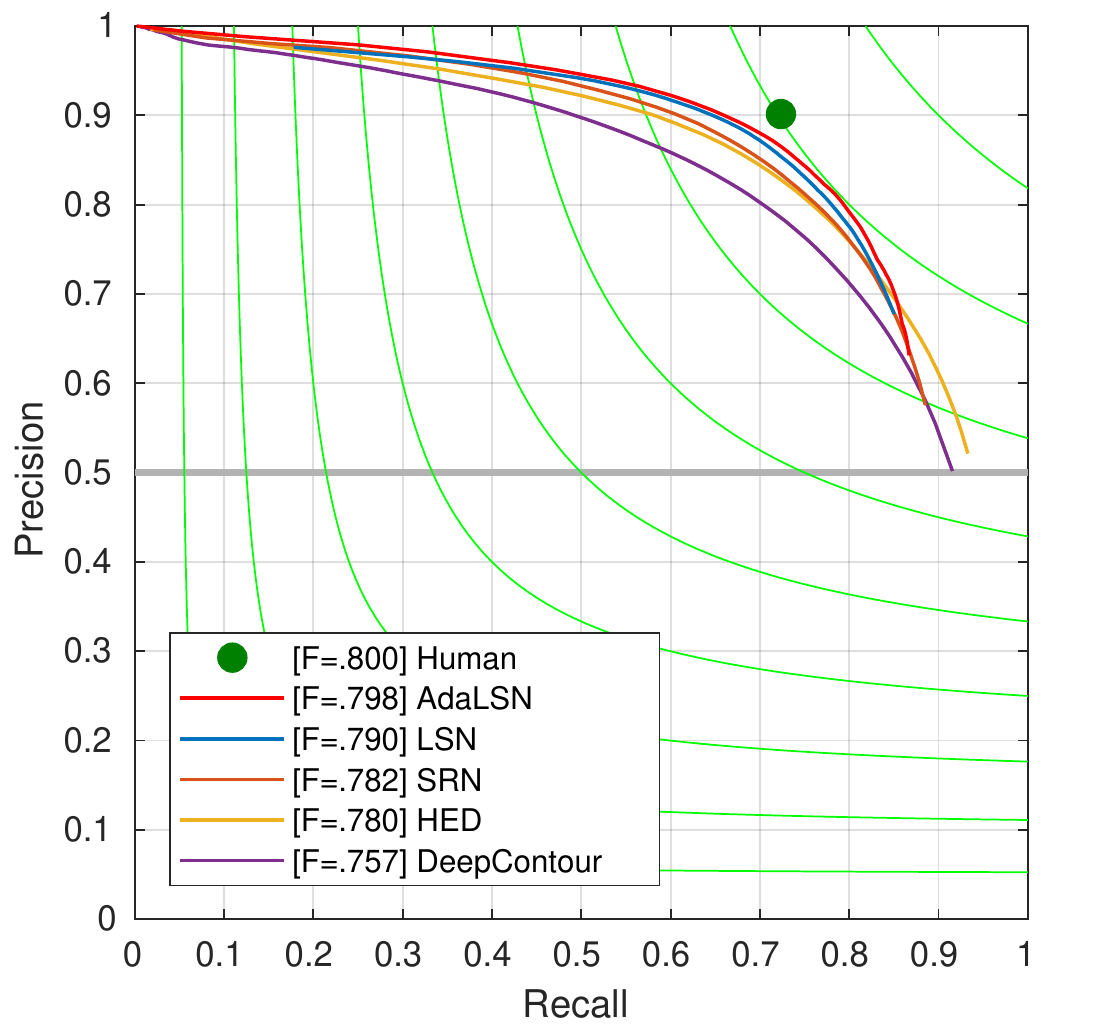}
\caption{The PR-curve on the BSDS500 edge detection dataset.}
\label{fig:EdgePR}
%\vspace{-0.2cm}
\end{figure}

\textbf{Edge Detection.} 
To demonstrate the general applicability of AdaLSN, we directly apply the model searched on the skeleton dataset to edge detection and report the edge detection performance on the BSDS500 dataset~\cite{BSDS500}, which is composed of 200 training images, 100 validation images, and 200 testing images. The F-score used as evaluation metrics is chosen an optimal scale for the entire dataset. As shown in Fig.~\ref{fig:EdgePR}, all the CNN-based approaches achieve good performance, which is comparable to human performance. AdaLSN reports the highest F-score of 0.798, which has a very small gap (0.002) to human performance. 
%The F-score with an optimal scale for the per image (OIS) was {\color{red}0.806}, which was even higher than human performance.

\begin{figure}[t]
\centering
\includegraphics[width=.5\textwidth]{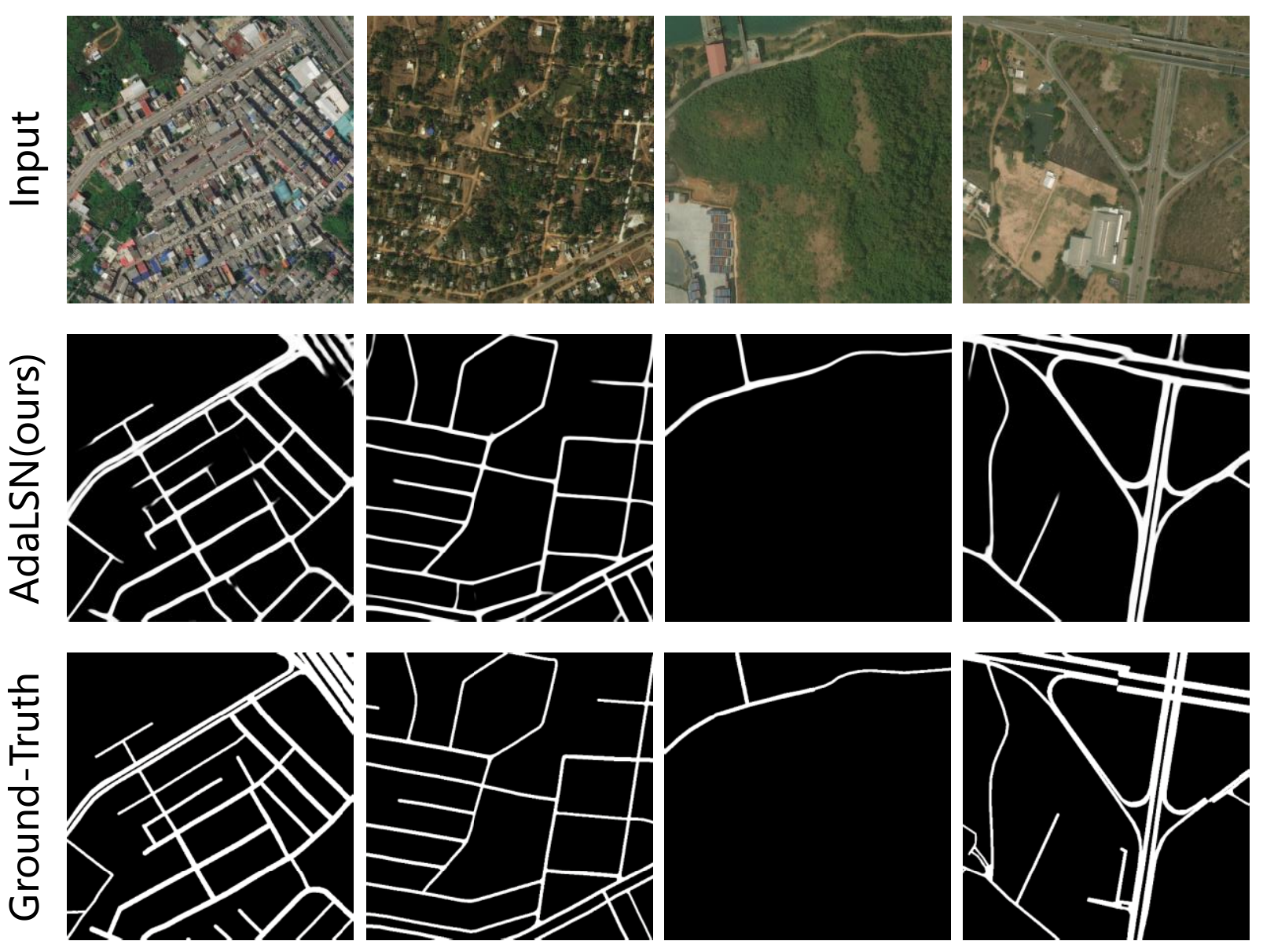}
\caption{Examples of the road extraction results of AdaLSN on the dataset for DeepGlobe Road Extraction Challenge~\cite{Deepglobe2018}.}
\label{fig:road}
%\vspace{-0.2cm}
\end{figure}

\textbf{Road Extraction.}
We also test our AdaLSN for road extraction in aerial images, using the dataset from the DeepGlobe Road Extraction Challenge~\cite{Deepglobe2018}. As the ground-truth annotations of the test dataset are not published, we randomly select 100 images from the training dataset as a test dataset and cut the rest images to $512\times512$ sub-images for training. Ada-LSN slightly outperforms DLinkNet~\cite{DLinkNet2018} (0.6354 vs 0.6321 mAP), which is one of the state-of-the-art approaches for road extraction. The detected road masks of our approach are shown in Fig.~\ref{fig:road}, which are very close to the ground-truth masks.

\begin{comment}
\begin{table}[t]
    \centering
    \caption{Road Extraction.}
    \label{tab.road}
    \begin{threeparttable}[b]
    \begin{tabular}{@{}cccc@{}}
    \toprule
          \textbf{Method}
        & \textbf{D-LinkNet34\cite{DLinkNet2018}}
        & \textbf{AdaLSN}
\\
        \midrule
         \textbf{IoU}
        & 0.6321
        & 0.6354
\\
    \bottomrule
    \end{tabular}
    \end{threeparttable}
\end{table}
\end{comment}

\section{Conclusion}
Object skeleton detection is a representative image-to-mask task in computer vision yet remains challenged by objects in different granularity. In this paper, we proposed adaptive linear span network (AdaLSN), with the aim to automatically configure and integrate scale-aware features for object skeleton detection. Following the linear span theory and driven by neural architecture search (NAS), AdaLSN configured complementary features for subspace and sum-space expansion. It thus optimized the multi-scale feature integration in a data-adaptive fashion. The significant higher accuracy compared with the state-of-the-arts and the general applicability to image-to-mask tasks demonstrated the effectiveness of the proposed AdaLSN approach. The NAS-driven feature space expansion provides a fresh insight for feature representation learning.

\ifCLASSOPTIONcaptionsoff
  \newpage
\fi

\section*{Acknowledgment}
The authors would like to express their sincere appreciation to the editors and the reviewers for their constructive comments. This work was supported in part by National Natural Science Foundation of China (NSFC) under Grant 61836012 and 61771447.

%%%%%%%%%%%%%%%%%%%%%%%%%%%%%%%%%%%%%%
%references
\bibliographystyle{IEEEtran}
\bibliography{IEEEabrv,egbib}
%\bibliography{IEEEabrv,refs}
%%%%%%%%%%%%%%%%%%%%%%%%%%%%%%%%%%%%%%

% that's all folks
\end{document}